\documentclass[11pt]{article}
\usepackage[margin=1in]{geometry}
\usepackage{amsmath,amssymb,amsthm,mathtools}
\usepackage[authoryear,round]{natbib}
\usepackage{graphicx,booktabs,multirow}
\usepackage[hidelinks]{hyperref}
\usepackage[capitalize,nameinlink]{cleveref}
\usepackage{enumitem}
\usepackage{microtype}
\setlist{nosep,leftmargin=*}

\newtheorem{theorem}{Theorem}
\newtheorem{lemma}{Lemma}
\newtheorem{proposition}{Proposition}
\newtheorem{corollary}{Corollary}
\theoremstyle{definition}
\newtheorem{assumption}{Assumption}
\newtheorem{definition}{Definition}
\theoremstyle{remark}
\newtheorem{remark}{Remark}

\renewcommand{\Pr}{\mathbb{P}}
\newcommand{\E}{\mathbb{E}}
\newcommand{\1}{\mathbf{1}}
\newcommand{\epsmin}{\varepsilon_{\min}}
\newcommand{\hlow}{h_{\mathrm{low}}}
\newcommand{\Grel}{\mathcal{G}^{\mathrm{rel}}}
\newcommand{\cA}{\mathcal{A}}
\newcommand{\cF}{\mathcal{F}}
\newcommand{\cH}{\mathcal{H}}
\DeclareMathOperator*{\essinf}{ess\,inf}

\title{Prevalence Bounds for Edited Watermarked Text:\\[2pt]
\large Power Envelopes and Calibration Sensitivity}
\author{Kuancheng Ye\\[2pt]
\small Department of Statistics, Donald Bren School of Information and Computer Sciences\\
\small University of California, Irvine\\
\small \texttt{kuanchey@uci.edu}}
\date{\today}

\begin{document}
\maketitle

\begin{abstract}
Watermark detectors decide whether one document came from a watermarked language model. Auditors often need a different number: the share of a corpus that did. We study this share for Gumbel-max and green-list (KGW) watermarks when every document may have been edited by an unknown amount. Without restrictions on the edits, the upper end of the identified set is $1$. Combining a known least-favourable next-token distribution with a sequential coupling, we derive a power envelope indexed by three budgets: a retention floor, an entropy budget, and a budget for in-place substitutions, which for Gumbel-max make edited positions look \emph{less} watermarked than human text. The envelope gives an identified set that is sharp for the alternatives it dominates, finite-sample bounds on the realized share of a fixed corpus with either the exact null or a null calibrated on an independent human sample, and confidence sequences for streams. In Gumbel-max simulations where the envelope binds, the bounds cover, and the threshold that maximizes the gap between envelope and null is generally as tight as twelve thresholds. On text from OPT-1.3B and Qwen3-1.7B, 12{,}000 fresh human documents scored under three keys show no robust departure from the exact null, and all bounds covered in resampling trials from finite pools of documents with random edits, substitutions and paraphrases; at a share of $0.2$ the exact-null bound is about $[0.14,0.38]$. These results are fragile in two ways we make precise. LLM-paraphrased KGW pools fall below the envelope, yet at 2000 documents the bounds still covered in these trials. A simulated null error of $0.01$, comparable to the largest deviation we observed at our thresholds, is likewise harmless at 2000 documents but breaks exact-null coverage at 100{,}000; a null calibrated on an independent human sample keeps coverage there, but needs a large reference sample to stay narrow.
\end{abstract}

\section{Introduction}\label{sec:intro}

Providers that watermark the output of their language models \citep{kirchenbauer2023watermark,aaronson2023watermark,kuditipudi2024robust,dathathri2024synthid} can test whether one text carries the watermark, and the statistical theory of such tests is well developed \citep{li2025framework,li2024robust,huang2026anytime}. Providers and the auditors they work with also face an aggregate question: what share of a collection of documents, such as submissions to a venue or posts on a platform, came from the watermarked model after users edited it? We call this the \emph{watermark-origin share}. Two features of it shape everything below. It concerns one provider's model, not machine-written text in general, because text from other models looks human to this test. And only the key holder can compute the statistics, so the analysis is run by the provider or by an auditor it gives access to.

The detection rate does not estimate this share. Human documents produce false positives, and watermarked documents that were edited, short or low in entropy are missed, by an amount that depends on how much each user edited. Correcting for the null rate, as Storey's estimator does \citep{storey2002direct}, removes the first error but not the second, and no point estimate says how wrong it might be. The share is only partially identified \citep{manski2003partial}. Because human documents' $p$-values are uniform under a well-calibrated key, the corpus is a two-groups mixture with a known null; without restricting the alternative, the upper end of the identified set is $1$ \citep{genovese2004stochastic,meinshausen2006estimating}.

We restrict the alternative through the watermark mechanism and state the restriction as three budgets a user must choose: a floor $\epsmin$ on the share of positions that keep their watermark signal, a cap $\sigma_{\max}$ on in-place substitutions, and a cap $\delta$ on next-token confidence. The resulting bounds are indexed by these budgets. Their lower end needs only a calibrated null. Their upper end needs the budgets, which cannot be estimated from the audited corpus, so we recommend reporting it over a range of budgets.

\paragraph{Contributions.}
\begin{enumerate}
\item \textbf{A power envelope derived from the mechanism} (\cref{sec:identification}). \Citet{li2025framework} showed that next-token distributions with $\max_wP(w)\le1-\delta$ have a two-point least-favourable member. A sequential coupling lifts it to edited documents of length $n$. We show that an in-place substitution leaves a key draw under which the replacement lost, so for Gumbel-max it makes edited text look less watermarked than human text (\cref{lem:subst}); the usual edit model, in which edited positions are exactly null, misses this. For the green-list scheme the same distribution is least favourable and the envelope has a closed form (\cref{prop:kgw}). The envelope gives a sharp identified set for the alternatives it dominates (\cref{thm:idset}); point-mass next-token distributions leave the share unidentified, and a reference sample identifies only $n$ moments $\pi\E[\varepsilon^k]$ (\cref{sec:negative}).
\item \textbf{Bounds on the realized share} (\cref{sec:inference}). For a fixed corpus we bound the realized share with fixed, heterogeneous labels, under a calibration condition that holds jointly given the design, with either an exact null or a null calibrated on an independent human reference sample (\cref{prop:realized,prop:calibrated}). Under i.i.d.\ sampling we cover the whole identified set (\cref{cor:band}), and for streams with drifting adoption we give a confidence sequence (\cref{thm:running}).
\item \textbf{Finite-sample behaviour and calibration sensitivity} (\cref{sec:sims,sec:experiments}). We study the bounds where the envelope binds, compare them fairly with single-threshold corrections, and examine null calibration with real models, several keys and independent human samples, and in a controlled simulation. In Gumbel-max simulations where the envelope binds, the single threshold that maximizes the gap between envelope and null is generally as tight as twelve thresholds, and both are much tighter than a detection count with worst-case sensitivity when power is low. With 12{,}000 fresh human documents and three keys we find no robust departure from the exact null, all bounds covered in resampling trials from finite pools of real-model text, and an earlier undercoverage did not replicate on independent text. Simulations show why this agreement should not be overread: the bound's sampling slack shrinks with corpus size, while an error in the null or a violation of the envelope does not, so either can be hidden at 2000 documents; for null errors we show that they are exposed at 100{,}000.
\end{enumerate}

\paragraph{Related work.}
\citet{han2026privacy} argue, on privacy grounds, that watermarks should target aggregate usage rather than individuals. They design a sequence-level watermark with a randomized-response structure, in which each text contributes one bit, and give a clipped moment estimator of the corpus fraction with bias, variance and mean-squared-error bounds. Their scheme is sensitive to edits, which they leave to future work, and they give no intervals. Our results apply to the token-level schemes already deployed and allow every document to be edited. \citet{li2025hybrid} estimate the proportion $\varepsilon$ of watermarked tokens \emph{within one} mixed human--AI text, show that $\varepsilon$ is identified for continuous pivots and not identified for the green-list pivot, and derive estimators that attain minimax lower bounds. Their results concern one document and do not include intervals. \citet{blanchet2026localization} localize the token positions within a document where the watermark survives. \citet{chai2026sample} give matching sample-complexity bounds for estimating the Gumbel-max watermark proportion within one document, under the same non-degeneracy condition $\max_wP(w)\le1-\delta$ that we use. \citet{huang2026anytime}, \citet{su2026anytime} and \citet{luo2026raoblackwell} build anytime-valid tests from e-values and e-processes that accumulate evidence token by token within one text. Robustness of detection to paraphrasing and mixing has been studied empirically \citep{krishna2023paraphrasing,kirchenbauer2024reliability,piet2023markmywords}. \citet{sander2024radioactive,sander2025contamination} aggregate watermark evidence over many texts to test whether a model was trained on watermarked outputs or whether a benchmark leaked into training, a different aggregate question from ours. Our unit of analysis is the document, our target is a corpus-level fraction, and our sequential procedures run across documents. They can take any of these per-document statistics as input. \citet{liang2024monitoring,liang2024mapping} estimate the share of AI-modified text in corpora without watermarks, using a word-frequency mixture whose null must be estimated. Correcting a raw positive rate for imperfect sensitivity and specificity is classical in epidemiology \citep{rogan1978estimating}, and bounds under misclassification or contamination are standard in econometrics \citep{horowitz1995identification,molinari2008partial}. Our upper bound is such a correction with a worst-case sensitivity that the watermark mechanism supplies. The exact null available with watermarks is what lets us state finite-sample guarantees. On the statistical side we use the two-groups model \citep{storey2002direct,genovese2004stochastic,meinshausen2006estimating}, partial identification \citep{manski2003partial,imbens2004confidence}, and anytime-valid inference \citep{ville1939etude,howard2021time,ramdas2023gametheoretic,waudbysmith2024estimating}. None of these tools is new here, and neither is the least-favourable NTP. The new part is the document-level envelope under unknown per-document edits, which converts the watermark mechanism and two interpretable budgets $(\epsmin,\delta)$ into a restriction on the alternative.

\section{Setup}\label{sec:setup}

\paragraph{Watermarks and pivots.}
A language model with a finite vocabulary $V$, $|V|\ge2$, produces a next-token distribution (NTP) $P_t$ at each position $t$. A watermark replaces ordinary sampling with a rule that uses pseudo-random numbers $\zeta_t$ derived from a secret key and the preceding $m$ tokens. We consider two schemes.
\begin{itemize}
\item \emph{Gumbel-max} \citep{aaronson2023watermark}. $\zeta_t=(U_{t,w})_{w\in V}$ with independent $U(0,1)$ entries, the emitted token is $w_t=\arg\max_w U_{t,w}^{1/P_t(w)}$, and the pivot is $Y_t=U_{t,w_t}$. The rule samples exactly from $P_t$.
\item \emph{Green list} \citep{kirchenbauer2023watermark}. $\zeta_t$ is a uniformly random subset $\Gamma_t\subset V$ of size $\gamma|V|\in\{1,\dots,|V|-1\}$, the logits of tokens in $\Gamma_t$ are raised by $b>0$, and the pivot is $Y_t=\1\{w_t\in\Gamma_t\}$. We call this scheme KGW.
\end{itemize}
For a document we score the positions whose $m$-token context has not already occurred earlier in the same document, and we write $n$ for the number of scored positions. Repeated contexts reuse the same $\zeta_t$ and would otherwise break independence \citep{fernandez2023three}.

\paragraph{Human documents.} Let $F_0$ denote $U(0,1)$ for Gumbel-max and $\mathrm{Bern}(\gamma)$ for KGW.

\begin{assumption}[Exact null]\label{ass:null}
For a document not generated by the watermarked model, the scored pivots are independent with distribution $F_0$, conditionally on the document's length and on everything observed before it.
\end{assumption}

\Cref{ass:null} is the reason watermark detection has exact error control. It holds for ideal key draws that are independent of human text and fresh for every scored context, which deduplication within a document only approximates. Fixed keys and shared contexts across documents can break it in principle, so we state it as an assumption and check it empirically in \cref{sec:experiments}.

\paragraph{Watermarked documents with unknown edits.}
Users edit generated text by different amounts, and edits act on pivots in three ways. A scored position whose token and context are both original keeps its dependence on the key. A position whose context was changed, by a deletion, an insertion or an earlier substitution, gets a new key draw; we assume such draws are previously unexposed and independent of the token, so its pivot is null. A position whose context is intact but whose token was replaced, by a substitution or an insertion, is different: its key draw is the one under which the \emph{original} token won, so the new token is one that lost, and its pivot is smaller than a null pivot (\cref{lem:subst}). Extending the hybrid-text model of \citet{li2024robust,li2025hybrid}, which has only the first two kinds, we let each watermarked document $i$ carry its own retention rate $\varepsilon_i$ and substitution rate $\sigma_i$, with $\varepsilon_i+\sigma_i\le1$. Conditionally on the history, on $(\varepsilon_i,\sigma_i)$ and on $P_{it}$,
\begin{equation}\label{eq:modelstar}
Y_{it}\;\sim\;(1-\varepsilon_i-\sigma_i)\,F_0+\varepsilon_i\,F_{1,P_{it}}+\sigma_i\,F_{s,it},
\end{equation}
where $F_{1,P}$ is the pivot distribution of the unedited scheme under NTP $P$ and $F_{s,it}$ is the pivot distribution at a substituted position. Two conditions are built into \eqref{eq:modelstar}. First, the replacement token may depend on the key draw $\zeta_{it}$ only through the emitted token: given the emitted token and the history, it is independent of $\zeta_{it}$. This holds when the editor uses the text up to that position and independent randomness; it can fail if the same context recurs later in the document and the editor reads the continuation. Second, which positions are retained, reset or substituted is decided without regard to the token at that position, beyond the history. An editor who keeps likely tokens and rewrites unlikely ones violates this, and can push documents below the envelope; neither condition follows from merely not holding the key. Equation \eqref{eq:modelstar} also describes a \emph{stochastic} edit process: given the past, each position is retained, reset or substituted with probabilities $\varepsilon_i$ and $\sigma_i$. An editor who fixes the realized fractions in advance is not covered. For two positions with $P=(\frac12,\frac12)$, retaining exactly one gives a smaller upper-tail probability $\Pr(T_2\ge5)$ ($0.0674$) than the envelope with $\epsmin=\frac12$ ($0.0709$). A new context that was also used elsewhere under the same key, for example in text that was later deleted, may violate this assumption, and an insertion counts as a substitution only when the inserted token differs from the token originally emitted there. With $\sigma_i\equiv0$ this is the model of \citet{li2024robust,li2025hybrid}. For Gumbel-max a direct calculation \citep{li2025framework} gives
\begin{equation}\label{eq:F1}
F_{1,P}(x)=\sum_{w\in V}P(w)\,x^{1/P(w)},\qquad x\in[0,1].
\end{equation}
The rates $(\varepsilon_i,\sigma_i)$ are unknown, heterogeneous, and may depend on the document. This is the main difference from the prevalence estimator of \citet{han2026privacy}, which corresponds to $\varepsilon_i\equiv1$ and $\sigma_i\equiv0$.

\begin{definition}[Mechanism class]\label{def:class}
For $\epsmin\in(0,1]$, $\sigma_{\max}\in[0,1-\epsmin]$ and $\delta\in(0,\frac12]$, a watermarked document belongs to $\cA(\epsmin,\delta)$ if $\varepsilon_i\ge\epsmin$, if $\sigma_i\le\sigma_{\max}$, if $\max_wP_{it}(w)\le1-\delta$ at every scored position, and if each $P_{it}$ is determined before $\zeta_{it}$ is drawn, with $\zeta_{it}$ fresh given the past. We suppress $\sigma_{\max}$ from the notation; $\sigma_{\max}=0$ recovers the model without substitutions.
\end{definition}

The budgets have plain meanings. $\epsmin$ is the smallest fraction of watermark signal one is willing to assume survives editing. $\sigma_{\max}$ is the largest fraction of positions one allows to be replaced in place. $\delta$ bounds how confident the model may be at any one position. \Cref{sec:negative} shows that neither budget can be dropped. \Cref{sec:experiments} shows that the per-position bound on $P_{it}$ is too strong for real text, while the document-level consequence that the identification result uses is not.

Lengths need separate care. Generation stops at an end-of-text token that is itself sampled with the key, and deduplication depends on which tokens were produced, so a document's scored length can carry information about its pivots. A stopping rule that ends a document after one position exactly when its first pivot is small would make short watermarked documents look human, even though every position satisfies \eqref{eq:modelstar}. We therefore state the length-conditional version as an assumption.

\begin{assumption}[Length-conditional calibration]\label{ass:length}
For a watermarked document in $\cA(\epsmin,\delta)$, \eqref{eq:modelstar} and the constraints of \cref{def:class} hold at every scored position also conditionally on the document's scored length $n_i$.
\end{assumption}

\Cref{ass:length} holds by construction when the scored positions and their number are chosen without reference to the key randomness, as in our simulations. For generated text it is an assumption. Its consequence, that the length-conditional alternative dominates the envelope, is what the per-stratum class check of \cref{sec:experiments} examines.

\paragraph{Document $p$-values and the two-groups model.}
For Gumbel-max we use $T_n=\sum_{t\le n}-\log(1-Y_t)$, which is $\Gamma(n,1)$ under the null, and $p=\bar\Gamma_n(T_n)$, where $\bar\Gamma_n$ is the upper tail of that law. For KGW we use $T_n=\sum_tY_t$ and the randomized binomial $p$-value $p=\bar B_n(T_n)+U\,b_n(T_n)$ with $U\sim U(0,1)$ independent, where $b_n$ and $\bar B_n$ are the probability mass function and strict upper tail of $\mathrm{Bin}(n,\gamma)$. Under \cref{ass:null} both $p$-values are exactly $U(0,1)$.

Let $Z_i\in\{0,1\}$ indicate a watermarked document. If documents $(Z_i,n_i,p_i)$ are independent and identically distributed, then within length $n$
\begin{equation}\label{eq:twogroups}
F_n(t):=\Pr(p_i\le t\mid n_i=n)=(1-\pi_n)\,t+\pi_n\,G_n(t),
\end{equation}
where $\pi_n=\Pr(Z_i=1\mid n_i=n)$ and $G_n(t)=\Pr(p_i\le t\mid Z_i=1,n_i=n)$ is the unknown alternative, a mixture over edits and NTPs. Lengths are observed, so the corpus prevalence is $\pi=\sum_n\Pr(n_i=n)\,\pi_n$. We condition on length because watermarked and human documents generally have different length distributions. Averaging over lengths first would mix alternatives with very different power. \paragraph{Targets.} Three quantities are easily conflated (\cref{tab:targets}). The main target is the \emph{realized share} $\rho$ of watermark-origin documents among the $N$ eligible documents of a corpus, those with at least $50$ scored positions. A document is watermark-origin if the provider's watermarked model generated it, however it was edited afterwards, including when another model rewrote it. Our guarantees cover only edits of the form \eqref{eq:modelstar}, so a rewritten document keeps its label but may fall outside the class. Documents stitched from several sources have no single source label and are outside our scope. Because short or heavily looped documents are excluded, $\rho$ can differ from the share among all submitted documents. Under i.i.d.\ sampling one can instead target $\pi_{\mathrm{cond}}=\sum_s(N_s/N)\pi_s$, the source prevalence given the observed stratum sizes. For a stream, the target is the running realized share $\bar\pi_k=\frac1k\sum_{i\le k}Z_i$, which requires no stationarity.

\begin{table}[t]
\centering
\caption{Targets, what each requires, and what it supports. All three rely on the power envelope with budgets $(\epsmin,\sigma_{\max},\delta)$ and on \cref{ass:length}.}
\label{tab:targets}
\small\setlength{\tabcolsep}{4pt}
\begin{tabular}{p{2.6cm}p{3.6cm}p{3.6cm}p{3.6cm}}
\toprule
target & additional assumptions & result & supports a statement like \\
\midrule
realized share $\rho$ of a fixed corpus & \cref{ass:calib}, jointly given labels and strata; exact null or a reference band & \cref{prop:realized,prop:calibrated} & ``between 12\% and 31\% of these documents came from our model'' \\
$\pi_{\mathrm{cond}}$ & i.i.d.\ sampling & \cref{cor:band} & a statement about the source population \\
running share $\bar\pi_k$ & each label chosen before its document is seen & \cref{thm:running} & monitoring with drift, at any stopping time \\
\bottomrule
\end{tabular}
\end{table}

\section{Identification}\label{sec:identification}

Without restrictions on $G_n$ in \eqref{eq:twogroups}, the identified set for $\pi_n$ is $[\pi_{n,L},1]$ with $\pi_{n,L}=1-\essinf_tf_n(t)$, where $f_n$ is the density of $F_n$ \citep{genovese2004stochastic}. The upper end is $1$ because a watermarked document that was edited heavily enough looks exactly like a human one. This section derives a restriction on $G_n$ from the mechanism class $\cA(\epsmin,\delta)$.

\subsection{Gumbel-max: a least-favourable NTP and a power envelope}

\begin{lemma}[Least-favourable NTP; \citealp{li2025framework}, Theorem~3.2 and Lemmas~3.2--3.4]\label{lem:lf}
Fix $x\in(0,1)$ and $\delta\in(0,\frac12]$. Over all distributions $P$ on $V$ with $\max_wP(w)\le1-\delta$,
\[
\sup_P F_{1,P}(x)=F^*(x;\delta):=(1-\delta)\,x^{1/(1-\delta)}+\delta\,x^{1/\delta},
\]
attained by the two-point distribution $P^*=(1-\delta,\delta,0,\dots,0)$.
\end{lemma}

\Citet{li2025framework} prove that $P\mapsto F_{1,P}(x)$ is convex and that the vertices of the constraint polytope are the permutations of $P^*$; they use this to find the optimal score function for per-document detection. For a general cap $c=1-\delta$ (feasible when $|V|c\ge1$) the maximizer has $\lfloor1/c\rfloor$ coordinates equal to $c$ and one coordinate carrying the remainder. We include a short proof in \cref{app:proofs} for completeness. Without substitutions ($\sigma_{\max}=0$), the lemma implies that every pivot of a document in $\cA(\epsmin,\delta)$ has conditional distribution function at most
\begin{equation}\label{eq:Fbar}
\bar F(x):=(1-\epsmin)\,x+\epsmin\,F^*(x;\delta)<x,\qquad x\in(0,1),
\end{equation}
because the right side of \eqref{eq:modelstar} decreases in $\varepsilon_i$ and increases in $F_{1,P}$, and $F_{1,P}(x)\le x$.

\begin{lemma}[Substituted positions]\label{lem:subst}
Let the emitted token $w$ be replaced by a token $w'\neq w$ that, given $w$ and the history, is independent of the key draw at that position, with $q=P(w')$. For Gumbel-max the replacement's pivot has distribution function
\[
F_{\mathrm{sub}}(x;q)=x^{1/q}+\frac{x\,(1-x^{1/q-1})}{1-q}\ \ge\ x,
\]
with limit $x(1-\log x)$ as $q\to1$. It is nondecreasing in $q$, so if $\max_wP(w)\le1-\delta$, then $F_{s}\le F_{\mathrm{sub}}(\cdot;1-\delta)$. For KGW the replacement is green with probability at least $g_{\mathrm{sub}}(q)=g/\{g+(1-g)(1+cq)\}$, where $g=(\gamma|V|-1)/(|V|-1)$ and $c=e^b-1$, and $g_{\mathrm{sub}}(q)\ge g_{\mathrm{sub}}(1-\delta)$.
\end{lemma}

The Gumbel part follows from a representation of the winning draw. Given that $w$ won, $a=U_w^{1/P(w)}$ is uniform and every other $U_v$ is uniform on $[0,a^{P(v)}]$, so the replacement's pivot is $a^qV$ with $a,V$ independent uniforms (\cref{app:proofs}). A likely replacement for an unlikely emitted token is the worst case. At $\delta=0.2$, the bound $F_{\mathrm{sub}}(\cdot;0.8)$ corresponds to a mean score of $0.40$ against $1$ for a null position, and the KGW floor is $0.14$ against $0.5$ ($\gamma=0.5$, $b=2$, large vocabulary). These are conservative bounds, not attained worst cases. For Gumbel-max, at a position selected for substitution independently of the emitted token, the mean score is at least $\frac12$: the conditional mean score $M(q)$ of the replacement is decreasing and convex in $q$ with $M(\frac12)=\frac12$, and a replacement chosen from the text has $\E[P(w')]\le\sum_wP(w)\max_{v\ne w}P(v)\le\frac12$. For Gumbel-max, substitutions therefore always subtract signal, since $F_{\mathrm{sub}}\ge x$. For KGW the floor lies below $\gamma$ but a particular substitution can also be greener than null, so the lemma gives a worst case, not a direction. With substitutions allowed, every pivot of a document in $\cA(\epsmin,\delta)$ has conditional distribution function at most
\begin{equation}\label{eq:Fbarsub}
\bar F(x)=(1-\epsmin-\sigma_{\max})\,x+\epsmin\,F^*(x;\delta)+\sigma_{\max}F_{\mathrm{sub}}(x;1-\delta),
\end{equation}
because the right side of \eqref{eq:modelstar} decreases in $\varepsilon_i$ and increases in $\sigma_i$. With $\sigma_{\max}=0$ this is \eqref{eq:Fbar}. For $\bar F(x)<x$ on all of $(0,1)$ the substitution budget must be small relative to the retention floor. Writing $q=1-\delta$, $A=x-x^{1/q}$ and $B=x-x^{1/\delta}$, one has $F_{\mathrm{sub}}(x;q)-x=qA/\delta$ and $\{x-F^*(x;\delta)\}/\{F_{\mathrm{sub}}(x;q)-x\}=\delta+(\delta^2/q)B/A$. For $\delta<\frac12$, $B>A$ on $(0,1)$ and $B/A\to1$ as $x\to0$, so $\bar F(x)<x$ on all of $(0,1)$ if and only if $\sigma_{\max}\le\epsmin\,\delta/(1-\delta)$: $\sigma_{\max}\le\epsmin/9$ at $\delta=0.1$ and $\sigma_{\max}\le\epsmin/4$ at $\delta=0.2$. At $\delta=\frac12$ the inequality must be strict.

\begin{lemma}[Sequential coupling]\label{lem:coupling}
Let $\phi$ be nondecreasing and $T_n=\sum_{t\le n}\phi(Y_t)$. If, for every $t$, the conditional distribution function of $Y_t$ given the past is pointwise at most $\bar F$, then $T_n$ stochastically dominates $\bar T_n=\sum_{t\le n}\phi(\bar Y_t)$ with $\bar Y_t$ independent with distribution $\bar F$.
\end{lemma}

\begin{proposition}[Power envelope]\label{prop:envelope}
Let $\bar T_n=\sum_{t\le n}-\log(1-\bar Y_t)$ and define
\[
h(t;n,\epsmin,\delta):=\Pr\big(\bar T_n\ge\bar\Gamma_n^{-1}(t)\big).
\]
For a watermarked document in $\cA(\epsmin,\delta)$ with a fixed number $n\ge1$ of scored positions, and for one with $n_i=n$ under \cref{ass:length}, and every $t\in(0,1)$,
\[
\Pr(p\le t)\ \ge\ h(t;n,\epsmin,\delta),
\]
where $\bar Y_t$ are independent with distribution function \eqref{eq:Fbarsub}. If $\bar F(x)<x$ on $(0,1)$, which for $\delta<\frac12$ holds exactly when $\sigma_{\max}\le\epsmin\delta/(1-\delta)$, and always when $\sigma_{\max}=0$, then $h(t;n,\epsmin,\delta)>t$ for all $t\in(0,1)$. With $\sigma_{\max}=0$, equality holds for the least-favourable mechanism $\varepsilon_i\equiv\epsmin$, $P_{it}\equiv P^*$.
\end{proposition}

The envelope depends on $(n,\epsmin,\delta)$ jointly. This answers the main objection to treating prevalence as a generic mixture problem: the width of the identified set is not set by $\epsmin$ alone but by how the mechanism converts retention, entropy and length into power (\cref{fig:grid}). Strict inequality $h>t$ needs a separate argument, because a coupling that is strict at each position does not by itself make the tail probability strictly larger at every threshold (\cref{app:proofs}).

\subsection{Sharp identified set}

Write $h_n(t)=h(t;n,\epsmin,\delta)$ and let $\Grel_n=\{G:\ G\text{ a distribution function on }[0,1],\ G(t)\ge h_n(t)\ \forall t\in(0,1)\}$ be the class of alternatives that dominate the envelope. By \cref{prop:envelope} and \cref{ass:length}, $\Grel_n$ contains every alternative generated by mixtures of documents in $\cA(\epsmin,\delta)$.

\begin{theorem}[Identified set]\label{thm:idset}
Suppose $h_n(t)>t$ on $(0,1)$, $F_n$ is absolutely continuous, and $F_n=(1-\pi_n)U+\pi_nG_n$ for some $\pi_n\in[0,1]$ and $G_n\in\Grel_n$. Then the identified set for $\pi_n$ is
\[
\Big[\pi_{n,L},\ \pi_{n,U}\Big],\qquad
\pi_{n,L}=1-\essinf_tf_n(t),\qquad
\pi_{n,U}=\min\Big\{1,\ \inf_{t\in(0,1)}\frac{F_n(t)-t}{h_n(t)-t}\Big\}.
\]
\end{theorem}

\begin{proof}[Proof sketch]
Validity: $f_n\ge1-\pi_n$ gives the lower end, and $F_n(t)-t=\pi_n(G_n(t)-t)\ge\pi_n(h_n(t)-t)$ gives the upper end. Sharpness: for any $\pi$ in the interval, $G^\pi(t)=t+(F_n(t)-t)/\pi$ is a distribution function exactly when $\pi\ge\pi_{n,L}$, and it lies in $\Grel_n$ exactly when $\pi\le\pi_{n,U}$. When $F_n\neq U$ we have $\pi_{n,L}>0$, so $\pi>0$ throughout; when $F_n=U$ the set is $\{0\}$.
\end{proof}

If a large substitution budget makes $h_n(t)\le t$ for some $t$, the infimum restricted to thresholds with $h_n(t)>t$ is still a valid upper end, but we do not claim sharpness. The Storey-type quantity $\sup_t(F_n(t)-t)/(1-t)$ is always a valid lower bound for $\pi_{n,L}$. It equals $\pi_{n,L}$ when $f_n$ is nonincreasing, because $(1-F_n(t))/(1-t)$ is the average of $f_n$ over $[t,1]$. Our inference procedures use this lower bound, which is conservative in general.

The sharpness is with respect to $\Grel_n$; the class actually generated by $\cA(\epsmin,\delta)$ is smaller, and \cref{app:relaxed} discusses how much is lost.

\begin{remark}[What the theorem uses]\label{rem:chain}
The argument runs through a chain of sufficient conditions: $\max_wP_t(w)\le1-\delta$ at each position implies $F_{1,P_t}\le F^*$ pointwise (\cref{lem:lf}), which implies $G_n\ge h_n$ (\cref{lem:coupling}). \Cref{thm:idset} uses only the last link. The first link is easy to interpret and can be checked from model logits, but it is much stronger than necessary. A distribution with top probability $0.9$ and the remaining mass spread over nine tokens violates the first two links. Its occasional low-probability tokens produce pivots near $1$, which makes its documents more detectable than under $P^*$ at the thresholds that matter in practice, and the bound in \cref{sec:sims} still covers. It does not dominate everywhere, though. At $x=0.01$, $F_{1,P}(x)\approx0.0054$ exceeds $F^*(x;0.2)\approx0.0025$, so for short documents the envelope can fail at thresholds near $1$. In our simulations, coverage failed only when a high top probability came with a remainder concentrated on few tokens, that is, nearly deterministic text. The theorem also relies on the retention and substitution constraints, fresh key draws, the edit model and length calibration; the entropy cap alone does not imply the document-level condition. \Cref{sec:experiments} finds that real text violates the per-position condition often, while the document-level check flagged only one stratum, by a small margin, under random edits or substitutions; the data cannot certify the condition.
\end{remark}

\paragraph{Length strata.}
In practice we pool lengths into strata and use, for every document in stratum $s$, the envelope $h^{(s)}(t)=\min_{n\in s}h(t;n)$ over the finitely many lengths the stratum can contain (all our real-data documents have at most $400$ scored positions). We have not proved that $h(t;n)$ is monotone in $n$, which would make this minimum the value at the smallest length. Numerically, for every $n$ from $50$ to $400$ on our threshold grid, at $\epsmin=0.3$ and $\delta\in\{0.1,0.2\}$ and for both schemes, the computed envelope is nondecreasing in $n$, so the minimum is attained at the smallest length of each stratum. The simulations use a single length. Weighting the stratum intervals by the observed stratum shares gives an identified set for the length-weighted prevalence $\sum_s(N_s/N)\pi_s$; the population prevalence uses the population stratum shares instead.

\subsection{Green-list watermarks}

For KGW, write $S=\sum_{w\in\Gamma}P(w)$ for the green mass and $c=e^b-1$. The green probability is $\varphi(S)=e^bS/(1+cS)$, so the excess over the null is $\varphi(S)-S=\psi(S):=cS(1-S)/(1+cS)$. Since $\E_\Gamma S=\gamma$, the expected green probability under $P$ is $\mu(P)=\gamma+\E_\Gamma\psi(S(P))$.

\begin{lemma}\label{lem:kgw}
$\psi$ is strictly concave on $[0,1]$, with $\psi''(S)=-2c(c+1)/(1+cS)^3$. Consequently $P\mapsto\mu(P)-\gamma$ is concave, its minimum over $\{\max_wP(w)\le1-\delta\}$ is attained at a vertex, and for $\delta\le\frac12$
\[
\inf_{P}\big(\mu(P)-\gamma\big)=\Delta_{\mathrm{KGW}}(\delta):=p_{10}\big[\psi(\delta)+\psi(1-\delta)\big],\qquad p_{10}=\frac{\gamma|V|\,(|V|-\gamma|V|)}{|V|(|V|-1)},
\]
attained at the same $P^*=(1-\delta,\delta,0,\dots)$ as for Gumbel-max.
\end{lemma}

So Gumbel-max maximizes a convex function of $P$ (the null-side distribution function, \citealp{li2025framework}) and KGW minimizes a concave one (the green excess). Both optimizations end at the same two-point vertex. Here $p_{10}$ is the probability that a specified one of two given tokens is green and the other is not, which tends to $\gamma(1-\gamma)$ as $|V|$ grows.

\begin{proposition}[KGW envelope]\label{prop:kgw}
Assume each $P_t$ is fixed before $\Gamma_t$ is drawn and $\Gamma_t$ is uniform given the past and $P_t$. For a watermarked document in $\cA(\epsmin,\delta)$ with a fixed number $n$ of scored positions, or with $n_i=n$ under \cref{ass:length}, every scored position is green with conditional probability at least $\bar\mu=\gamma+\epsmin\Delta_{\mathrm{KGW}}(\delta)-\sigma_{\max}\{\gamma-g_{\mathrm{sub}}(1-\delta)\}$, and the randomized $p$-value satisfies
\[
\Pr(p\le t)\ \ge\ h_{\mathrm{KGW}}(t;n):=\sum_{k=0}^n\binom nk\bar\mu^k(1-\bar\mu)^{n-k}\,\min\Big\{1,\max\Big\{0,\frac{t-\bar B_n(k)}{b_n(k)}\Big\}\Big\}.
\]
\end{proposition}

Moreover $h_{\mathrm{KGW}}(t;n)>t$ for $t\in(0,1)$ whenever $\bar\mu>\gamma$: the randomized rejection probability is nondecreasing and nonconstant in the count, and $\mathrm{Bin}(n,\bar\mu)$ then strictly dominates the null count. If $\bar\mu\le\gamma$ the envelope carries no information and the upper end of the identified set is $1$. At $\epsmin=0.3$, $\delta=0.2$, this happens once $\sigma_{\max}\ge0.129$ ($\gamma=0.5$, $b=2$, large vocabulary). \Cref{thm:idset} therefore holds verbatim with $h_n$ replaced by $h_{\mathrm{KGW}}$. At $\epsmin=0.3$, $n=200$ and $t=0.05$, the KGW envelope ($\gamma=0.5$, $b=2$, $\sigma_{\max}=0$, large vocabulary) is $0.236$ for $\delta=0.1$ and $0.368$ for $\delta=0.2$. The conservative Gumbel envelope of \cref{sec:inference} is $0.260$ and $0.433$. The two schemes are of the same order, with KGW slightly weaker.

\subsection{Two negative results}\label{sec:negative}

\begin{proposition}[Degenerate NTPs]\label{prop:degenerate}
Consider documents of a fixed length $n$ without substitutions ($\sigma_{\max}=0$). If the class allows documents whose NTP is a point mass at every position, together with arbitrary mixtures of such documents and the other alternatives, then for both schemes such a document has an exactly uniform $p$-value, whatever its retention. The upper end of the identified set over this class is then $1$.
\end{proposition}

For Gumbel-max, $F_{1,P}(x)=x$ when $P$ is a point mass. For KGW, $S\in\{0,1\}$ almost surely and $\psi(S)=0$. With $\sigma_{\max}=0$, as $\delta\to0$ both envelopes tend to $t$ pointwise. Pointwise collapse alone does not force the identified upper end to $1$, because the infimum over $t$ can still use thresholds near $1$; the proposition concerns exact point masses. For KGW, $\Delta_{\mathrm{KGW}}$ falls from $0.154$ at $\delta=0.2$ to $0.017$ at $\delta=0.01$ (large vocabulary). This is a different failure from the within-document non-identifiability of \citet{li2025hybrid}, which comes from the green-list pivot being binary with an unknown green probability, and which they avoid for Gumbel-max by assuming non-singular NTPs. Here both schemes lose identification once point-mass NTPs are allowed, consistent with that assumption. Low-entropy text limits what any corpus-level analysis can learn, however strong the retention assumption.

A natural alternative is to estimate the token-level watermark distribution $F_1$ from an unedited reference sample, as an analogue of the known response rates in \citet{han2026privacy}. The next result shows what that buys.

\begin{proposition}[Reference sample identifies finitely many moments]\label{prop:reference}
Suppose $F_1\neq F_0$ is known, pivots follow \eqref{eq:modelstar} with $F_{1,P}\equiv F_1$ and no substitutions ($\sigma_i\equiv0$), and $\varepsilon_i\sim\Lambda$ independently of everything else. With substitutions the observed law also involves $\pi\E[\sigma(F_s-F_0)]$ and the statement does not apply.
\begin{enumerate}[label=(\roman*)]
\item For $n=1$ the data identify exactly $\pi\,\E_\Lambda[\varepsilon]$.
\item For any $n$, if $\Lambda$ may put mass at $0$, then for every $\pi'\in[\pi,1]$ with $\pi'>0$ the pair $\big(\pi',\ \tfrac{\pi}{\pi'}\Lambda+(1-\tfrac{\pi}{\pi'})\delta_0\big)$ generates the same distribution of pivots as $(\pi,\Lambda)$. The upper end of the identified set is $1$.
\item If pivots are conditionally independent given $\varepsilon$, the data identify exactly $\{\pi\,\E_\Lambda[\varepsilon^k]\}_{k=1}^n$. A floor $\varepsilon\ge\epsmin$ then gives $\pi\le\pi\E_\Lambda[\varepsilon]/\epsmin$.
\end{enumerate}
\end{proposition}

A reference sample turns the problem into a truncated moment problem. It does not remove the need for a retention floor. Part (ii) holds only for the distribution of pivots, which is all the analyst observes once the text is reduced to watermark statistics.

\section{Bounds on the realized share}\label{sec:inference}

\subsection{A conservative envelope}
The bounds below need a function that lies below $h$, not merely an approximation to it. Write $X=-\log(1-\bar Y)$, whose distribution function $\bar F(1-e^{-s})$ is available in closed form from \eqref{eq:Fbarsub}. We replace $X$ by $X_{\mathrm{tr}}=\min\{\lfloor X/\Delta s\rfloor\Delta s,\ S_{\max}\}\le X$, compute the lattice probabilities of $X_{\mathrm{tr}}$ exactly from that closed form, convolve them $n$ times by an FFT long enough to hold the full support, and round the threshold $\bar\Gamma_n^{-1}(t)$ up to the lattice. In exact arithmetic each step moves the result downward. At $n=50,100,200$ (lattice step $0.02$) and $n=300$ (step $0.05$), the FFT tail probabilities agree with direct summation of nonnegative terms to within $10^{-13}$; longer documents in the simulations ($n$ up to $1000$) were not checked this way, and we subtract a tolerance of $10^{-8}$, clip to $[0,1]$, and call the result $\hlow(t;n)$. The floating-point error is measured, not bounded. At $n=100,200,300$, the envelope with $\Delta s=0.01$ is conservative by at most $0.024$ relative to a lattice four times finer. For KGW we evaluate the finite sum in \cref{prop:kgw}, with $\gamma(1-\gamma)$ in place of $p_{10}$.

\subsection{The realized share of a fixed corpus}
The target of an audit is the share of watermark-origin documents among the $N$ eligible documents actually collected, $\rho=N^{-1}\sum_iZ_i$. Labels need not be random. Let strata $s=1,\dots,K$ group documents by length, with $N_s$ documents and $K_s=\sum_{i\in s}Z_i$ watermarked ones, and fix thresholds $t_1<\dots<t_J$ in advance. Write $S_{sj}=\sum_{i\in s}\1\{p_i\le t_j\}$ and $h_s=\min_{n\in s}\hlow(\cdot;n)$.

\begin{assumption}[Conditional calibration]\label{ass:calib}
Let the design $\mathcal D$ consist of the labels $Z_1,\dots,Z_N$, the stratum memberships and the order in which documents are processed. Given $\mathcal D$ and the earlier $p$-values, a human document in stratum $s$ has $\Pr(p_i\le t)=H_s(t)$ and a watermarked one has $\Pr(p_i\le t)\ge h_s(t)$.
\end{assumption}

\Cref{ass:calib} is a joint condition on the whole design, not a document-by-document one. We impose this joint condition directly. A sufficient special case is that, conditionally on the labels and stratum memberships, documents are independent, human scores have distribution function $H_s$ ($H_s(t)=t$ under \cref{ass:null} and \cref{ass:length}), and watermarked scores dominate $h_s$. The document-level assumptions alone do not establish the joint condition, and it can fail even if every label is fixed before its own document is scored: if $Z_2=\1\{p_1\le1/2\}$, then given both labels $p_1$ is no longer uniform. Strata enter only through membership. The watermarked bound $h_s=\min_{n\in s}\hlow(\cdot;n)$ holds for any mixture of lengths within the stratum, and $H_s$ is the human law mixed over the stratum's lengths. We keep $H_s$ general because the null can fail in practice (\cref{sec:experiments}).

\begin{proposition}[Realized share, exact null]\label{prop:realized}
Under \cref{ass:calib} with $H_s(t)=t$, let $\beta_s=\{N_s\log(2KJ/\alpha)/2\}^{1/2}$. With probability at least $1-\alpha$, simultaneously for all $s$,
\[
\max_j\frac{S_{sj}-\beta_s-N_st_j}{1-t_j}\ \le\ K_s\ \le\ \min_{j:\,h_s(t_j)>t_j}\frac{S_{sj}+\beta_s-N_st_j}{h_s(t_j)-t_j},
\]
and summing the clipped stratum bounds over $s$ and dividing by $N$ bounds $\rho$.
\end{proposition}

\begin{proof}
Condition on $\mathcal D$, so that $N_s$ and $K_s$ are fixed. Order the documents as in $\mathcal D$, let $\mathcal F_i=\sigma(\mathcal D,p_1,\dots,p_i)$ and $m_{ij}=\Pr(p_i\le t_j\mid\mathcal F_{i-1})$. Then $\1\{p_i\le t_j\}-m_{ij}$ is a martingale difference taking values in an interval of length $1$, so by the Azuma--Hoeffding inequality and a union bound over the $KJ$ pairs, $|S_{sj}-\sum_{i\in s}m_{ij}|\le\beta_s$ for all $(s,j)$ with probability at least $1-\alpha$. Since $m_{ij}=t_j$ for human documents and $m_{ij}\in[h_s(t_j),1]$ for watermarked ones, $\sum_{i\in s}m_{ij}$ lies between $N_st_j+K_s(h_s(t_j)-t_j)$ and $N_st_j+K_s(1-t_j)$; rearranging gives the display.
\end{proof}

Thresholds with $h_s(t_j)\le t_j$ are skipped; if none remains, the upper bound is $K_s\le N_s$. Stratum bounds are clipped to $[0,N_s]$, and an empty stratum contributes nothing. If a lower bound exceeds the upper bound, the procedure returns an empty confidence set, which indicates incompatibility with the assumptions at the stated confidence level. This is the fixed-$N$, multi-threshold form of the martingale argument behind \cref{thm:running}, with a union bound in place of a time-uniform boundary. It allows fixed, heterogeneous and non-i.i.d.\ labels, but only through \cref{ass:calib}, which must hold jointly given the design. Labels that adapt to earlier $p$-values are covered by \cref{thm:running}, not by this proposition.

\subsection{A calibrated null}
When human scores under a fixed key are not exactly uniform, an independent reference sample of human documents scored under the same key can stand in for the known null. Suppose that with probability at least $1-\alpha_{\mathrm{ref}}$ we have $H_{L,s}\le H_s\le H_{U,s}$ at every threshold and stratum. A DKW band per stratum with a Bonferroni split over strata gives this when the reference documents of stratum $s$ are i.i.d.\ draws from $H_s$, independent of the audited corpus. A stratum without reference documents gets the trivial band $[0,1]$. The lower bound below uses only thresholds with $H_{U,s}(t_j)<1$, the upper bound only those with $h_s(t_j)>H_{U,s}(t_j)$, and otherwise the rules after \cref{prop:realized} apply.

\begin{proposition}[Realized share, calibrated null]\label{prop:calibrated}
Under \cref{ass:calib}, with $\beta_s$ as above, with probability at least $1-\alpha-\alpha_{\mathrm{ref}}$, simultaneously for all $s$,
\[
\max_{j:\,H_{U,s}(t_j)<1}\frac{S_{sj}-\beta_s-N_sH_{U,s}(t_j)}{1-H_{U,s}(t_j)}\ \le\ K_s\ \le\ \min_{j:\,h_s(t_j)>H_{U,s}(t_j)}\ \max_{H\in\{H_{L,s}(t_j),H_{U,s}(t_j)\}}\frac{S_{sj}+\beta_s-N_sH}{h_s(t_j)-H}.
\]
\end{proposition}

\begin{proof}
On both events, $\sum_{i\in s}m_{ij}=N_sH+K_s(\bar G-H)$ with $H=H_s(t_j)\in[H_{L,s},H_{U,s}]$ and $\bar G\in[h_s(t_j),1]$. The lower bound follows from $K_s(1-H)\ge S_{sj}-\beta_s-N_sH$ and the fact that $(S-\beta-N_sH)/(1-H)$ is nonincreasing in $H$ when $S-\beta\le N_s$. For the upper bound, $K_s\le(S+\beta-N_sH)/(h-H)$ for the true $H$, and this function of $H$ is monotone on $[H_{L,s},H_{U,s}]$ because its derivative has the sign of $S+\beta-N_sh$; its maximum is at an endpoint.
\end{proof}

The price is a transport assumption: the reference documents must have the same null law as the audited human documents, which requires the same key, the same scoring and comparable text. The lower end needs only the null band, so it is the more robust half of the interval.

\subsection{The identified set under i.i.d.\ sampling}
If documents are independent and identically distributed, the identified set of \cref{thm:idset} can be estimated directly.

Let $\hat F_s$ be the empirical distribution function of the $N_s$ $p$-values in stratum $s=1,\dots,K$, let $\hlow^{(s)}(t)=\min_{n\in s}\hlow(t;n)$ be the conservative envelope minimized over the stratum's possible lengths, and let $\mathcal T\subset(0,1)$ be a finite grid. Set $b_s=\sqrt{\log(2K/\alpha)/(2N_s)}$ and
\[
L_s=\max\Big\{0,\ \max_{t\in\mathcal T}\frac{\hat F_s(t)-b_s-t}{1-t}\Big\},\qquad
U_s=\min\Big\{1,\ \min_{t\in\mathcal T:\ \hlow^{(s)}(t)>t}\frac{\hat F_s(t)+b_s-t}{\hlow^{(s)}(t)-t}\Big\},
\]
with $U_s=1$ when no grid point has $\hlow^{(s)}(t)>t$. Every stratum is assumed nonempty.

$L_s$ is the DKW-based lower confidence bound for the non-null proportion of \citet[Theorem~3.1]{genovese2004stochastic}, extended by \citet{meinshausen2006estimating}; $U_s$ is new and uses the envelope.

\begin{corollary}[Band inversion]\label{cor:band}
Suppose that, conditionally on the stratum memberships, the $p$-values within each stratum are independent and identically distributed according to \eqref{eq:twogroups}, with alternatives in the relaxed classes defined by $\hlow^{(s)}$. Then with probability at least $1-\alpha$, simultaneously for all $s$, the identified set of $\pi_s$ is contained in $[L_s,U_s]$. In particular $\sum_sw_sL_s\le\pi_{\mathrm{cond}}\le\sum_sw_sU_s$ with $w_s=N_s/N$, where $\pi_{\mathrm{cond}}=\sum_s(N_s/N)\,\pi_s$ is the prevalence conditional on the observed stratum sizes. It is neither the population prevalence nor the realized fraction of watermarked documents.
\end{corollary}

\begin{proof}
By the Dvoretzky--Kiefer--Wolfowitz inequality with Massart's constant and a union bound over strata, $\sup_t|\hat F_s(t)-F_s(t)|\le b_s$ for all $s$ with probability at least $1-\alpha$. On that event the Storey-type quantity at each $t$ is at least $(\hat F_s(t)-b_s-t)/(1-t)$, so $L_s\le\pi_{s,L}$. For the upper end, $\pi_{s,U}\le(F_s(t)-t)/(\hlow^{(s)}(t)-t)\le(\hat F_s(t)+b_s-t)/(\hlow^{(s)}(t)-t)$ at each grid point where $\hlow^{(s)}(t)>t$. Restricting to a finite grid only enlarges the interval.
\end{proof}

The corollary covers the whole identified set of $\pi_{\mathrm{cond}}$, the prevalence conditional on the observed stratum sizes; replacing $t$ by a calibrated band $[H_L,H_U]$ as in \cref{prop:calibrated} gives the corresponding calibrated version. Plugging $\hat F_s$ into \cref{thm:idset} without the band does not give usable point estimates of the endpoints, because the infimum picks up downward noise where $h(t)-t$ is small.

\subsection{Streams}\label{sec:cs}
Documents may arrive one at a time, with adoption that drifts. Within a stratum fix $t$ and let $B_i=\1\{p_i\le t\}$. 
We target $\bar\pi_k=\frac1k\sum_{i\le k}Z_i$ and drop the assumption on $\Pr(Z_i=1\mid\cdot)$. The labels may follow any sequence, chosen adaptively, provided that $Z_i$ is fixed before document $i$ is seen. Formally, let $\cH_{i-1}=\sigma(p_1,\dots,p_{i-1},Z_1,\dots,Z_i)$ and assume the calibration statements above hold conditionally on $\cH_{i-1}$, with the same caveat on $\hlow$. Then
\begin{equation}\label{eq:mbounds2}
t+Z_i\,a\ \le\ m_i:=\E[B_i\mid\cH_{i-1}]\ \le\ t+Z_i(1-t),\qquad a=\hlow(t)-t>0.
\end{equation}
$S_k=\sum_{i\le k}(B_i-m_i)$ is a martingale. Given $\cH_{i-1}$, its increment lies in $[-m_i,1-m_i]$, an interval of length $1$, so by Hoeffding's lemma its conditional moment generating function is at most $e^{\lambda^2/8}$. Mixing $\exp(\lambda S_k-\lambda^2k/8)$ over $\lambda\sim N(0,1/\rho)$, with $\rho>0$ fixed before the stream starts, \citep{robbins1970statistical,howard2021time} gives the boundary
\[
\mathrm{bd}_k=\sqrt{2(k/4+\rho)\,\log\!\Big(\tfrac1\alpha\sqrt{(k/4+\rho)/\rho}\Big)}.
\]

\begin{theorem}[Confidence sequence for the realized fraction]\label{thm:running}
Let $\tilde S_k=\sum_{i\le k}(B_i-t)$. Under the conditions above,
\[
\Pr\Big(\forall k:\ \frac{\tilde S_k-\mathrm{bd}_k}{k(1-t)}\ \le\ \bar\pi_k\ \le\ \frac{\tilde S_k+\mathrm{bd}_k}{k\,a}\Big)\ \ge\ 1-\alpha .
\]
\end{theorem}

\begin{proof}
The mixture is a nonnegative supermartingale, so $|S_k|<\mathrm{bd}_k$ for all $k$ with probability at least $1-\alpha$. On that event, summing \eqref{eq:mbounds2} gives $a\sum_{i\le k}Z_i\le\tilde S_k-S_k<\tilde S_k+\mathrm{bd}_k$ and $(1-t)\sum_{i\le k}Z_i\ge\tilde S_k-S_k>\tilde S_k-\mathrm{bd}_k$.
\end{proof}

The target moves with $k$, so a plain running intersection is not valid. Earlier bounds can still be carried forward: $k\bar\pi_k\in[jL_j,\ jU_j+(k-j)]$ for all $j<k$ on the same event. The sub-Gaussian boundary ignores variance information and uses one threshold. In our simulations the sequence was wider than the betting one, although the two target different quantities under different assumptions. Empirical-Bernstein boundaries may tighten it. With documents of mixed lengths, the same argument bounds $\sum_iZ_ia_{n_i}$. Using $\min_ia_{n_i}$ gives a conservative bound on the count.

A betting confidence sequence for a fixed prevalence, with averaging over thresholds, is given in \cref{app:betting}.

\section{Simulations}\label{sec:sims}

All simulations draw Gumbel-max pivots directly from \eqref{eq:modelstar}. A watermarked token is the top token with probability $q_t$, and the remaining mass $1-q_t$ is spread over nine other tokens. Human documents have exactly uniform $p$-values. Unless stated otherwise, $\alpha=0.05$, the envelope is the conservative $\hlow$, and intervals come from \cref{cor:band} with a single length stratum.

\begin{figure}[t]
\centering
\includegraphics[width=\linewidth]{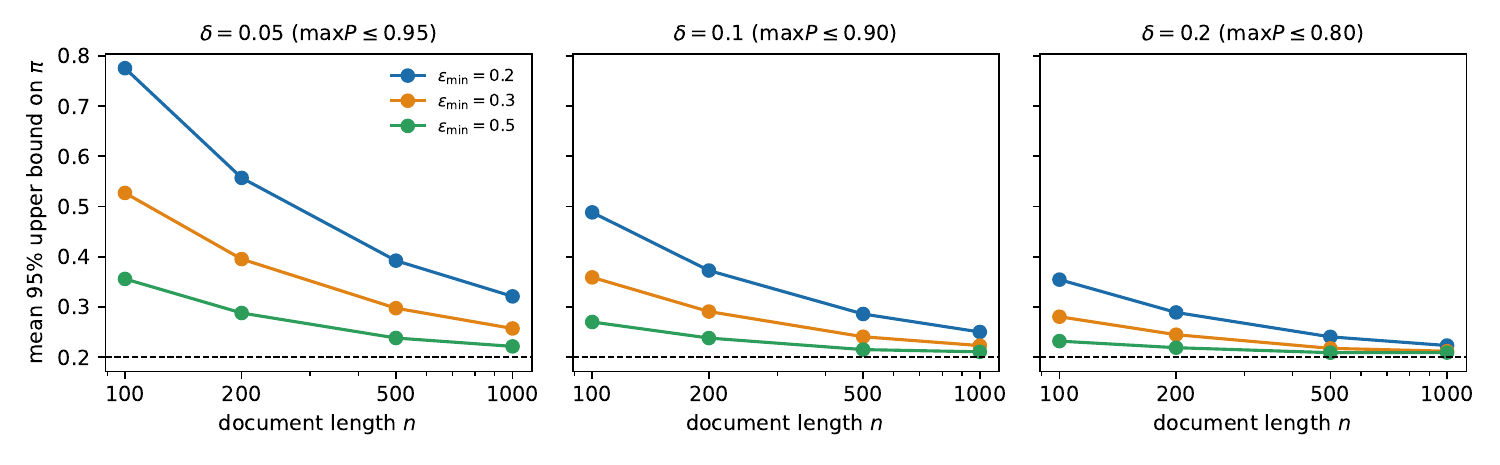}
\caption{Mean 95\% upper confidence bound on $\pi$ (true value $0.2$, dashed) from \cref{cor:band} with the conservative envelope, for one fixed in-class data-generating process ($N=20{,}000$ documents, 20 replications). Panels vary the assumed entropy budget $\delta$; lines vary the assumed retention floor $\epsmin$.}
\label{fig:grid}
\end{figure}

\paragraph{How wide is the identified set?}
\Cref{fig:grid} reports the mean upper confidence bound for $\pi=0.2$ on one fixed data-generating process, with $\varepsilon_i\sim U[0.5,1]$ and $q_t\sim U[0.3,0.8]$, which lies in $\cA(\epsmin,\delta)$ for every assumed $(\epsmin,\delta)$ in the figure. Only the assumptions change across panels. The lower bound sits near $0.192$ in every cell, because it does not use the envelope. The upper bound responds to all three quantities. At $n=200$ and $\epsmin=0.3$ it falls from $0.395$ to $0.291$ to $0.245$ as $\delta$ goes from $0.05$ to $0.1$ to $0.2$. At $\epsmin=0.3$ and $\delta=0.1$ it falls from $0.359$ at $n=100$ to $0.223$ at $n=1000$. The weakest cell ($n=100$, $\epsmin=0.2$, $\delta=0.05$) gives $0.776$, nearly uninformative. In 25 of the 36 cells the bound is below $0.3$. Coverage was at least $0.95$ in every cell. These numbers answer a practical question directly: how much survival and entropy one must be willing to assume to pin the prevalence down to a useful range.

\paragraph{Coverage.}
On a fixed process with $\varepsilon_i\sim U[0.3,1]$ and $q_t\sim U[0.3,0.8]$, we crossed three in-class assumptions $(\epsmin,\delta)\in\{(0.3,0.2),(0.3,0.1),(0.2,0.2)\}$ with $n\in\{200,500\}$, $\pi\in\{0.05,0.2\}$ and $N\in\{2000,20000\}$. Across these 24 cells, with 100 replications each, coverage of the true $\pi$ ranged from $0.93$ to $1.00$ with mean $0.97$. The band is simultaneous over $t$ and therefore conservative; the one cell at $0.93$ is within Monte Carlo error of $0.95$ for 100 replications. Typical intervals at $(\epsmin,\delta)=(0.3,0.2)$ were $[0.191,0.218]$ for $n=500$, $\pi=0.2$, $N=20{,}000$ and $[0.025,0.104]$ for $n=200$, $\pi=0.05$, $N=2000$.

\paragraph{When the assumption fails.}
\Cref{tab:outclass} holds $\varepsilon_i\equiv0.3$ and every NTP fixed, and compares assumptions that do and do not contain the truth. When the remainder sits on a single token, assuming $\delta=0.2$ (that is, $\max P\le0.8$) for text whose top probability is $0.85$ or $0.9$ makes the upper bound fall below the truth in every replication. When the same top probability comes with a remainder spread over nine tokens, the interval still covers, as explained in \cref{rem:chain}. The failures we observed involve nearly deterministic text, such as code, templates or lists, where the top token is likely and the alternatives are few.

\begin{table}[t]
\centering
\caption{Coverage under violated assumptions. $n=500$, $N=20{,}000$, $\pi=0.2$, $\varepsilon_i\equiv0.3$, 100 replications. The NTP at every position puts mass $q$ on the top token and splits $1-q$ over one or nine other tokens.}
\label{tab:outclass}
\begin{tabular}{ccccc}
\toprule
top prob.\ $q$ & remainder over & assumed $\delta$ & coverage & mean upper bound \\
\midrule
0.90 & 1 token  & 0.1 (holds)    & 1.00 & 0.220 \\
0.90 & 1 token  & 0.2 (violated) & 0.00 & 0.114 \\
0.85 & 1 token  & 0.1 (holds)    & 1.00 & 0.236 \\
0.85 & 1 token  & 0.2 (violated) & 0.00 & 0.182 \\
0.90 & 9 tokens & 0.2 (violated) & 0.99 & 0.217 \\
0.85 & 9 tokens & 0.2 (violated) & 0.97 & 0.219 \\
\bottomrule
\end{tabular}
\end{table}

\paragraph{Weak signal, fair comparison and the whole identified set.}
To probe the regime where the envelope binds, we simulate watermarked documents from the least-favourable mechanism itself ($\varepsilon_i\equiv0.3$, $P_t\equiv P^*$), whose $p$-value law equals the envelope, with lengths and entropy budgets chosen so that detection power at $t=0.05$ ranges from $0.19$ to $0.93$. We add a stress test in which a share $\sigma=0.07$ of positions draws its pivot from the substitution bound of \cref{lem:subst}, which no editor need attain, just inside the strictness boundary $\epsmin\delta/(1-\delta)=0.075$, and a mixed setting in which half the documents are strong. Corpora have $N\in\{500,2000\}$ documents and $\pi\in\{0.05,0.2\}$, with 200 replications. We compare five bounds on the realized share, all at level $0.05$: \cref{prop:realized} over twelve pre-specified thresholds; the same bound at the single threshold $t=0.05$, which is a detection count with worst-case sensitivity \citep{rogan1978estimating}; a single threshold chosen without data to maximize $h(t)-t$; a single threshold chosen to minimize $\sqrt{t(1-t)}/(h(t)-t)$; and a single threshold chosen in each replication on a fresh, independent pilot corpus from the same population.

All five covered the realized share in at least $99.5\%$ of replications in every setting. \Cref{tab:weak} shows the upper ends at $N=2000$. Against the fixed threshold $0.05$, every other choice is much tighter when power is low (at power $0.19$ and $\pi=0.2$, about $0.32$ against $0.45$) and about equal when power is high. At $N=2000$, the data-free rule that maximizes $h(t)-t$ was the tightest of the four choices in every setting but one, the mixed setting at $\pi=0.2$ ($0.311$ against $0.293$ for the pilot and $0.299$ for the multi-threshold bound). A pilot-tuned threshold was on average about as tight as the multi-threshold bound, but the pilot's choice varied widely between replications (from $t=0.1$ to $0.8$ in the low-power settings, and from $0.01$ to $0.8$ overall), so its upper end varied more. The rule minimizing $\sqrt{t(1-t)}/(h(t)-t)$ picked $t=0.01$ in most settings, where it was worse than the fixed threshold, and $0.1$ or $0.2$ in the low-power settings, where it was much tighter. So the gain over a detection count comes from the envelope and from a sensible threshold, and the multi-threshold bound mainly buys robustness to the choice: in these settings it was never far from the best single threshold, at the price of a union-bound penalty. These are Gumbel-max simulations; we did not repeat the comparison for KGW or real corpora.

Under i.i.d.\ labels, \cref{cor:band} covered the whole population identified set, computed from the known population distribution, in $93\%$ to $100\%$ of replications (the lowest is within Monte Carlo error of $95\%$ for 200 replications), with intervals narrower than those of \cref{prop:realized}, which pays for not assuming i.i.d.\ labels.

\begin{table}[t]
\centering
\caption{Average 95\% upper bound on the realized share, $N=2000$, 200 replications, least-favourable documents unless stated. ``$\max h-t$'' is the single threshold maximizing $h(t)-t$ (value in parentheses); ``pilot'' chooses one threshold per replication on a fresh independent pilot corpus (10th--90th percentile of the upper bound in brackets).}
\label{tab:weak}
\small\setlength{\tabcolsep}{3.5pt}
\begin{tabular}{lccccccc}
\toprule
setting & power at $0.05$ & $\pi$ & 12 thresholds & $t=0.05$ & $\max h-t$ & pilot \\
\midrule
$\delta=0.2$, $n=50$ & 0.19 & 0.05 & 0.166 & 0.281 & 0.151 (0.4) & 0.161 [0.11, 0.21] \\
 & & 0.20 & 0.331 & 0.446 & 0.316 (0.4) & 0.330 [0.28, 0.38] \\
$\delta=0.05$, $n=400$ & 0.25 & 0.05 & 0.159 & 0.232 & 0.146 (0.3) & 0.154 [0.11, 0.19] \\
 & & 0.20 & 0.321 & 0.410 & 0.310 (0.3) & 0.317 [0.28, 0.36] \\
$\delta=0.2$, $n=200$, $\sigma=0.07$ & 0.26 & 0.20 & 0.325 & 0.410 & 0.313 (0.3) & 0.320 [0.28, 0.36] \\
$\delta=0.2$, $n=200$ & 0.46 & 0.20 & 0.272 & 0.294 & 0.264 (0.2) & 0.269 [0.24, 0.30] \\
$\delta=0.2$, $n=400$ & 0.70 & 0.20 & 0.256 & 0.260 & 0.249 (0.2) & 0.254 [0.24, 0.27] \\
$\delta=0.2$, $n=800$ & 0.93 & 0.20 & 0.244 & 0.240 & 0.240 (0.05) & 0.244 [0.23, 0.26] \\
mixed, $\delta=0.2$, $n=200$ & 0.73 & 0.20 & 0.299 & 0.435 & 0.311 (0.2) & 0.293 [0.26, 0.33] \\
\bottomrule
\end{tabular}
\end{table}

\paragraph{Sensitivity to the null.}
The real-data results rest on human $p$-values being uniform, and our checks can only bound the departure. To see what a small departure does, we give human documents the distribution function $H(t)=t-4\eta\,t(1-t)$, whose largest deviation from uniform is $|\eta|$ at $t=0.5$; $\eta>0$ means too few small $p$-values, which biases the exact-null upper end down, and $\eta<0$ the reverse. Watermarked documents are least-favourable ($n=200$, $\delta=0.2$, $\varepsilon=0.3$) or, as a second alternative, half least-favourable and half strong. We compare \cref{prop:realized} with the exact null and \cref{prop:calibrated} with a DKW band from an independent human reference sample of $M\in\{4000,40{,}000\}$ documents, redrawn in every replication, with $\alpha$ split equally between the corpus and the reference. There are 200 replications for $N\le20{,}000$ and 50 for $N=100{,}000$.

\Cref{tab:nullsens} shows the exact-null coverage for least-favourable documents. At $N=2000$ a deviation of $0.01$, comparable to the largest deviation at our twelve thresholds in human samples of twelve thousand documents, costs nothing, and even $0.02$ leaves coverage at $0.96$ or more. At $N=100{,}000$ the same $0.01$ drops coverage to $0$ at $\pi=0.05$ and to $0.26$ at $\pi=0.2$, and $\eta=-0.01$ drops it to $0.10$ at $\pi=0.05$. The pattern follows the slack per document of \cref{prop:realized}, $\{\log(2J/\alpha)/(2N)\}^{1/2}$, which is $0.039$, $0.012$ and $0.006$ at the three corpus sizes. With half the watermarked documents strong, failures are milder at $\pi=0.2$ (coverage $0.94$ at $\eta=0.01$, $N=100{,}000$) but just as severe at $\pi=0.05$. The calibrated bound covered in every cell, for both alternatives and both reference sizes. Its price is width: at $N=100{,}000$, $\pi=0.05$ and $\eta=0$, the exact-null bound averages $[0.034,0.061]$ and the calibrated one $[0.022,0.074]$ with $M=40{,}000$ and $[0.004,0.095]$ with $M=4000$; at $\pi=0.2$ they are $[0.170,0.215]$, $[0.153,0.227]$ and $[0.129,0.247]$. To keep the calibrated bound close to the exact-null one, the reference band must be small relative to the corpus's own slack, and at $N=100{,}000$ even $40{,}000$ reference documents leave it noticeably wider.

\begin{table}[t]
\centering
\caption{Coverage of the realized share by \cref{prop:realized} with the exact null when human $p$-values have distribution function $t-4\eta t(1-t)$. Least-favourable watermarked documents ($n=200$, $\delta=0.2$, $\varepsilon=0.3$); 200 replications ($N\le20{,}000$) or 50 ($N=100{,}000$). The calibrated bound of \cref{prop:calibrated} covered in every cell.}
\label{tab:nullsens}
\small
\begin{tabular}{lcccccc}
\toprule
& \multicolumn{3}{c}{$\pi=0.05$} & \multicolumn{3}{c}{$\pi=0.2$} \\
\cmidrule(lr){2-4}\cmidrule(lr){5-7}
$\eta$ & $N=2000$ & $20{,}000$ & $100{,}000$ & $2000$ & $20{,}000$ & $100{,}000$ \\
\midrule
$0$ & 1.000 & 1.000 & 1.000 & 1.000 & 1.000 & 1.000 \\
$+0.005$ & 1.000 & 0.965 & 0.720 & 0.995 & 1.000 & 0.980 \\
$+0.010$ & 0.995 & 0.715 & 0.000 & 1.000 & 0.940 & 0.260 \\
$+0.020$ & 0.965 & 0.005 & 0.000 & 0.990 & 0.225 & 0.000 \\
$-0.005$ & 0.995 & 1.000 & 0.960 & 1.000 & 1.000 & 1.000 \\
$-0.010$ & 1.000 & 0.925 & 0.100 & 1.000 & 1.000 & 0.960 \\
$-0.020$ & 0.960 & 0.060 & 0.000 & 1.000 & 0.770 & 0.000 \\
\bottomrule
\end{tabular}
\end{table}

\section{Experiments with real models}\label{sec:experiments}

\paragraph{Generation.}
We watermark OPT-1.3B \citep{zhang2022opt} and Qwen3-1.7B \citep{yang2025qwen3} with both schemes at temperature 1, with key-derived randomness seeded by SHA-256 of the key and the previous $m=3$ tokens. KGW uses $\gamma=0.5$ and $b=2$. Prompts are the first 30 tokens of documents from the C4 validation split \citep{raffel2020exploring}. We generate up to 300 new tokens and keep documents with at least 50 new tokens, about 2000 per model and scheme. The analysis then drops documents with fewer than 50 \emph{scored} positions, leaving between 1862 and 2011 unedited watermarked documents per model and scheme. At every position we log the unwatermarked NTP's top probability $\max_wP_t(w)$, which is the quantity that \cref{def:class} constrains. Human documents in the original batch are 4000 further C4 documents, truncated to 330 tokens and scored from position 30, the same offset as the generated text. Scoring deduplicates repeated contexts within a document. This removes $4.4\%$ to $4.7\%$ of positions in the original human batch and $3.4\%$ to $7.6\%$ in KGW text, but $34.5\%$ (Qwen3) to $44.6\%$ (OPT) in Gumbel-max text. With a fixed key and context hashing, a recurring three-token context reuses the same key draw, so when the NTP is also similar the same token tends to follow, and the text can fall into a loop. Deduplication drops these repeated positions, so they carry no weight in the statistics below.

\paragraph{Perturbations.}
\emph{Random edits.} Each watermarked document draws a nominal retention $\varepsilon_i\sim U[\varepsilon_{\mathrm{lo}},1]$ with $\varepsilon_{\mathrm{lo}}\in\{0.3,0.5\}$. \emph{Scattered} edits replace each token independently with probability $1-\varepsilon_i^{1/(m+1)}$, so that a position and its context survive with probability about $\varepsilon_i$; \emph{span} edits replace contiguous spans of geometric length with mean 8. Replacement tokens are drawn from the pool of all tokens the same model generated, including the document's own. These edits are close to, but not exactly, \eqref{eq:modelstar}. Neighbouring positions share token survivals, and a unigram replacement at an intact context is a substitution in the sense of \cref{lem:subst}, although a mild one because such tokens have small probability under the NTP. The realized retention $\varepsilon_{\mathrm{eff}}$, the fraction of scored positions whose token and context are both intact, falls below the nominal value. With $\varepsilon_{\mathrm{lo}}=0.3$, between $4\%$ and $33\%$ of documents have $\varepsilon_{\mathrm{eff}}<0.3$ ($4\%$ for scattered edits of KGW text, $14\%$ to $33\%$ otherwise). A realized fraction below $0.3$ does not by itself violate a floor on the retention probability, but it shows that $\epsmin=0.3$ is not a conservative choice here.

\emph{Plausible substitutions.} To probe \cref{lem:subst} directly, each watermarked document draws a substitution rate $r_i\sim U[0.05,0.35]$. At each position whose last $m$ tokens are unchanged, the token is replaced with probability $r_i$ by a token drawn from the generating model's own NTP at that position, excluding the emitted token. The choice uses only the text. The realized share of substituted positions with intact context is $\sigma_{\mathrm{eff}}$, with mean $0.10$ to $0.12$ and 95th percentile $0.16$ to $0.17$ across pools; the mean realized retention is $0.43$ to $0.52$.

\emph{Low-power pools.} Two further pools move toward the regime where the envelope binds. \emph{Near-floor} edits draw $\varepsilon_i\sim U[0.3,0.4]$, giving mean realized retention $0.28$ to $0.35$ and detection rates of $0.88$ to $0.98$. \emph{Short} documents keep only the first 50 to 100 generated tokens before scattered edits with $\varepsilon_{\mathrm{lo}}=0.3$, giving a median of 72 to 74 scored positions among analysed documents and detection rates of $0.81$ to $0.99$.

\emph{Paraphrasing.} As an assumption-deviation experiment, we paraphrase 320 documents per model and scheme with Qwen2.5-3B-Instruct \citep{yang2024qwen25}, prompted to rewrite sentence by sentence at similar length (temperature $0.7$, top-$p$ $0.9$, at most 400 new tokens), and rescore them under the watermark tokenizer. Rewrites with fewer than 20 scored positions are dropped, and the analysis then keeps those with at least 50, leaving 274 to 311 per pool. Paraphrasing changes the NTPs and correlates edits across positions, so it is outside \eqref{eq:modelstar}.

\paragraph{Human documents and keys.}
Human text comes from two disjoint batches of C4 validation documents: an original batch of 4000 (eligible documents 2600--6599) and a fresh batch of 12000 (6600--18599). Both are scored under the original key. For Qwen3 with KGW and OPT with Gumbel-max, the fresh batch is also scored under two further keys, and 1000 watermarked documents with scattered edits are generated under each further key. Scoring the same texts under several keys separates variation due to the key from variation due to the text sample.

\paragraph{Design.}
Each corpus has $N=2000$ eligible documents, exactly $\pi N$ of them watermarked with $\pi\in\{0.05,0.2\}$, so the target is the realized share. Watermarked documents are drawn without replacement from a pool (with replacement only for the small paraphrase pools at $\pi=0.2$), and human documents without replacement from an 8000-document audit split of the fresh batch. The remaining 4000 human documents form the reference sample for \cref{prop:calibrated}; the reference and audit split is redrawn every 50 corpora. We use three strata of scored length, $[50,100)$, $[100,200)$ and $[200,\infty)$, the twelve thresholds $t\in\{0.01,0.02,0.05,0.1,0.2,\dots,0.9\}$, $\alpha=0.05$ for \cref{prop:realized} and $\alpha=\alpha_{\mathrm{ref}}=0.025$ for \cref{prop:calibrated}, $\epsmin=0.3$, $\delta\in\{0.1,0.2\}$, $\sigma_{\max}\in\{0,0.05\}$, and 200 corpora per cell. Results are conditional on the generated pools, which are finite samples of the populations of interest. Because corpora are drawn from fixed pools, the next human $p$-value given the earlier ones follows the remaining pool rather than $H_s$, and one reference band serves 50 corpora. The coverage rates below therefore describe the procedure on these pools; they are a diagnostic, not a test of the guarantees, which the simulations of \cref{sec:sims} check with everything redrawn.

\paragraph{Null calibration.}
In the fresh batch, deduplication removed $4.6\%$ to $4.8\%$ of positions. On the fresh batch, the eight combinations of model, scheme and key give Kolmogorov--Smirnov statistics $D$ of $0.006$ to $0.013$ ($n\approx12{,}000$), with uncorrected $p$-values from $0.03$ to $0.80$; one of the eight is below $0.05$, and none is significant after a Bonferroni correction for eight tests. The largest deviation of the empirical distribution function from uniform at the twelve thresholds is $1.4$ to $2.9$ binomial standard errors, at most $0.011$ in absolute terms. Earlier, on the original batch of 4000 documents, Qwen3 with KGW under the original key gave $D=0.025$ and $p=0.011$ (four tests, Bonferroni $p\approx0.045$), with the deviation concentrated around $p\approx0.6$. The same key on the fresh batch gives $p=0.50$, and the two batches differ from each other (two-sample KS $p=0.017$; the largest difference of empirical distribution functions is $0.023$, $2.7$ standard errors). For OPT with Gumbel-max the batches agree (two-sample $p=0.45$). So the earlier departure was not reproduced on independent text under the same key; whether it reflects sampling variation or a difference in the composition of the two batches we cannot tell. Three keys and two batches cannot rule out a key effect of the size we can no longer detect.

\paragraph{Coverage and width.}
\Cref{tab:realized} reports the realized-share bounds at $\pi=0.2$. Over all four model and scheme pairs, six pools, both prevalences, both $\delta$, and both substitution budgets, every bound covered the realized share in all 200 corpora: \cref{prop:realized} with the exact null, the same bound at the single threshold $0.05$, and \cref{prop:calibrated} with the reference-calibrated null. At a share of $0.2$ the exact-null bound is roughly $[0.14,0.38]$ under random edits and substitutions, which can rule out large shares but cannot pin the share near $0.2$. Calibrating the null raises the upper end by $0.04$ to $0.06$ and lowers the lower end by $0.02$ to $0.06$, and by up to $0.11$ for the short pool, whose reference strata are small. Under random edits, a detection count with worst-case sensitivity at $t=0.05$ gives upper ends of $0.56$ to $0.75$. A substitution budget of $\sigma_{\max}=0.05$ raises the upper end on the substitution pools from $0.38$--$0.40$ to $0.46$--$0.56$. Span edits (not shown) give exact-null endpoints within $0.012$ of scattered edits. The short pool gives the narrowest bounds only because its watermarked documents fall in a length stratum with few human documents, so length itself is informative there; length-matched corpora would not show this.

\begin{table}[t]
\centering
\caption{Realized-share bounds on real-model corpora, $N=2000$, true share $0.2$, $\delta=0.2$, $\sigma_{\max}=0$, 200 corpora. Entries are the average lower and upper ends of \cref{prop:realized} (exact null); in parentheses, the average upper end with the reference-calibrated null (\cref{prop:calibrated}) and with a single threshold $t=0.05$. Every bound in every cell covered the realized share in all 200 corpora.}
\label{tab:realized}
\small\setlength{\tabcolsep}{3pt}
\resizebox{\linewidth}{!}{\begin{tabular}{lcccc}
\toprule
pool & OPT / Gumbel & Qwen3 / Gumbel & OPT / KGW & Qwen3 / KGW \\
\midrule
random edits (scattered) & $[0.14,0.38]$ (0.42; 0.56) & $[0.14,0.37]$ (0.42; 0.64) & $[0.14,0.41]$ (0.45; 0.72) & $[0.15,0.38]$ (0.43; 0.75) \\
substitutions & $[0.13,0.38]$ (0.42; 0.57) & $[0.14,0.38]$ (0.43; 0.65) & $[0.13,0.40]$ (0.44; 0.72) & $[0.14,0.38]$ (0.43; 0.72) \\
near-floor & $[0.14,0.36]$ (0.41; 0.63) & $[0.14,0.36]$ (0.42; 0.65) & $[0.13,0.40]$ (0.44; 0.70) & $[0.14,0.38]$ (0.45; 0.68) \\
short & $[0.18,0.30]$ (0.35; 0.34) & $[0.17,0.30]$ (0.36; 0.35) & $[0.16,0.32]$ (0.36; 0.37) & $[0.16,0.30]$ (0.35; 0.36) \\
paraphrase & $[0.11,0.36]$ (0.40; 0.45) & $[0.09,0.36]$ (0.40; 0.45) & $[0.04,0.31]$ (0.35; 0.44) & $[0.02,0.27]$ (0.33; 0.34) \\
\bottomrule
\end{tabular}}
\end{table}

\paragraph{The earlier undercoverage.}
An earlier analysis, kept here because it shaped our conclusions, used a different design: i.i.d.\ labels, $N=4000$, human documents resampled with replacement from the original 4000-document batch, and the narrower DKW band of \cref{cor:band}. For Qwen3 with KGW at $\pi=0.05$ it covered in only $72\%$ of corpora. The same design with the fresh batch covers in $99.5\%$, and \cref{prop:realized} covers in $100\%$ with either batch; with \cref{prop:calibrated}, five independent reference splits of the fresh batch all gave $100\%$. The failure did not reproduce under the same key on the fresh batch, so it cannot be attributed to the key alone; an interaction between the key and the composition of the original batch is not ruled out. It was tied to the original human batch together with a design whose sampling slack is small. We read this as a warning, made precise in \cref{sec:sims}: a null error of the size seen in the original batch is absorbed by the slack of a 2000-document corpus but not by that of a much larger one.

\paragraph{Assumptions under paraphrase and at the token level.}
Coverage held for paraphrased corpora too, but that is not evidence that the envelope holds. The normalized check $r_s(t)=\{\hat G_s(t)-t\}/\{\hlow^{(s)}(t)-t\}$, which must be at least $1$ under the class, is below $1$ in every stratum of the KGW paraphrase pools and in three of the six Gumbel-max strata (down to $0.39$ for OPT with KGW; \cref{tab:s1}). For random edits, substitutions and the low-power pools it is $1.01$ to $1.23$ in every stratum except one, the middle stratum of the Qwen3 Gumbel-max near-floor pool ($0.90$, 153 documents). These minima sit at $t\approx0.94$--$0.97$ and are nearly the same across pools, so for in-class pools the check is driven by the envelope near $t=1$ and is only weakly informative. The paraphrase pools differ by scheme. For KGW they fall below the envelope also at thresholds up to $0.5$, by up to $0.18$ in the empirical rejection rate (Qwen3, 157 documents, about $4.6$ binomial standard errors). For Gumbel-max the shortfalls occur only at $t>0.9$, where the check is weak, and the gap is positive for $t\le0.5$. Coverage at $N=2000$ therefore does not validate the envelope; violations that persist in the population can matter as the sampling slack shrinks. We did not test larger $N$, and a violation at one stratum and threshold can be offset by slack elsewhere in the final bound. Separately, $10\%$ to $19\%$ of scored positions have top probability above $0.9$, so no useful $\delta$ satisfies \cref{def:class} position by position, as \citet{li2025framework} also observed; the document-level condition is the one to state.

\begin{table}[t]
\centering
\caption{Per-token versus document-level conditions. Left: share of scored (deduplicated) positions whose unwatermarked top probability exceeds $0.9$ and $0.99$, and the share of generated positions removed by deduplication. Right: smallest normalized ratio $\min_{s,t}r_s(t)$ for the paraphrased pools ($\epsmin=0.3$, $\delta=0.2$); values below $1$ mean the empirical pool falls below the envelope at that stratum and threshold, with no allowance for sampling error in the pool.}
\label{tab:s1}
\small\setlength{\tabcolsep}{4pt}
\begin{tabular}{lcccc}
\toprule
& \multicolumn{2}{c}{$\max_wP_t(w)>$} & & paraphrased \\
\cmidrule(lr){2-3}
model / scheme & 0.9 & 0.99 & dedup removed & min ratio \\
\midrule
OPT / Gumbel   & 0.111 & 0.050 & 0.446 & 0.763 \\
Qwen3 / Gumbel & 0.188 & 0.094 & 0.345 & 0.836 \\
OPT / KGW      & 0.098 & 0.041 & 0.034 & 0.392 \\
Qwen3 / KGW    & 0.177 & 0.085 & 0.076 & 0.620 \\
\bottomrule
\end{tabular}
\end{table}

\begin{figure}[t]
\centering
\includegraphics[width=\linewidth]{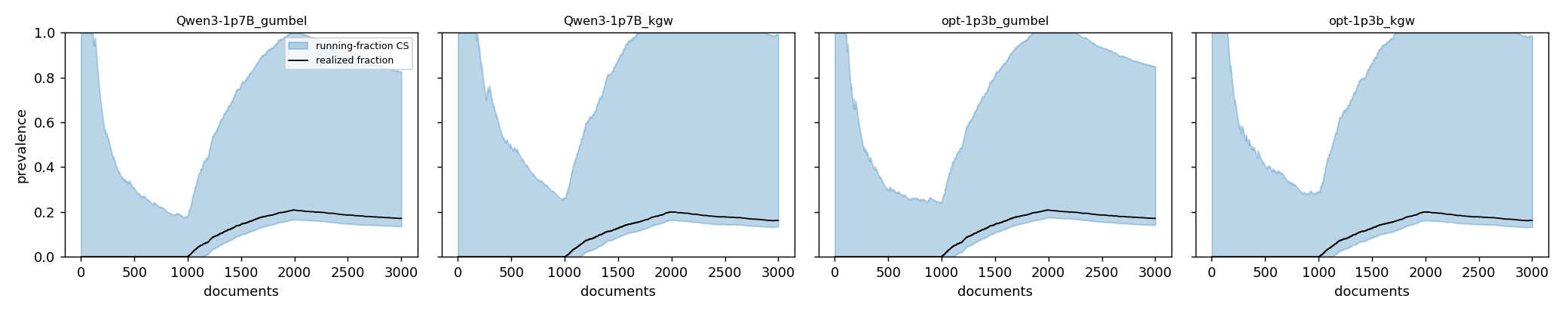}
\caption{Realized-fraction confidence sequence (\cref{thm:running}, shaded) and the realized fraction $\bar\pi_k$ (black) on a stream whose prevalence moves from $0$ to $0.4$ to $0.1$, for two models and two schemes. Scattered edits with $\varepsilon_{\mathrm{lo}}=0.5$; assumed $(\epsmin,\delta)=(0.3,0.2)$, $t=0.05$, envelope at $n=100$.}
\label{fig:seq}
\end{figure}

\paragraph{Streams.}
We feed 3000 documents of length at least 100 in three phases of 1000, with prevalence $0$, then $0.4$, then $0.1$, using scattered edits for the watermarked documents. \Cref{fig:seq} shows the realized-fraction sequence of \cref{thm:running} at $t=0.05$ for $\varepsilon_{\mathrm{lo}}=0.5$. It covered $\bar\pi_k$ at all 3000 times in all eight runs. Using the envelope at $n=100$ for longer documents relies on the numerical monotonicity in $n$ reported in \cref{sec:identification}. The lower end tracks the realized fraction closely. The upper end is loose at $t=0.05$, where $a=\hlow(t)-t$ is only about $0.2$. In a rerun with $\varepsilon_{\mathrm{lo}}=0.3$ it ended at $0.83$ to $1.00$ for a realized fraction of $0.17$. Choosing $t=0.3$ instead raises $a$ to $0.36$--$0.40$, and on the same stream the final upper end falls to $0.36$--$0.41$, again with coverage at all times. After the jump from $0$ to $0.4$ the upper end rises, since this sequence does not intersect over time. That is the price of validity under arbitrary drift.

\section{Discussion}\label{sec:discussion}

The watermark-origin share of a corpus is not identified without assumptions on how much signal survives editing, and no assumption on survival identifies it if next-token distributions may be point masses; nearly deterministic text makes the signal weak. With both, the watermark mechanism gives a power envelope, in closed form for KGW and computed conservatively for Gumbel-max, and the envelope gives bounds indexed by three budgets $(\epsmin,\sigma_{\max},\delta)$. The lower end needs only a calibrated null. The upper end needs the budgets, which cannot be estimated from the audited corpus.

\paragraph{What the slack hides.}
Take one stratum $s$ and one threshold $t$, and let $\rho_s=K_s/N_s$. The unclipped exact-null upper end of \cref{prop:realized}, $U_{s,t}=\{S_s(t)+\beta_s-N_st\}/\{N_s(h_s(t)-t)\}$, satisfies
\[
U_{s,t}-\rho_s=\frac{\xi_s(t)+b_s-(1-\rho_s)\,d_s(t)+\rho_s\,r_s(t)}{h_s(t)-t},
\]
where $b_s=\beta_s/N_s$ is the slack, $\xi_s(t)=\{S_s(t)-\sum_{i\in s}m_i(t)\}/N_s$ the martingale noise, $d_s(t)=t-H_s(t)$ the null error and $r_s(t)=\bar G_s(t)-h_s(t)$ the margin of the alternative, with $\bar G_s$ the average conditional rejection probability of the watermarked documents. The upper end fails when a null error $d_s>0$ or a class violation ($r_s<0$) outweighs the slack and the margin; symmetrically, $d_s<0$, too many small human $p$-values, can push the lower end above $\rho_s$. The slack shrinks like $N^{-1/2}$; null errors and class violations do not shrink. Two of our findings are instances of this. KGW paraphrase pools fall below the envelope, yet the bounds covered at $N=2000$. A simulated null error comparable to the largest deviation at our thresholds in human samples is harmless at $N=2000$ but breaks coverage at $N=100{,}000$ (\cref{sec:sims}). Coverage in moderate corpora is therefore weak evidence that the assumptions hold, and more precise bounds hide less. For large corpora we recommend the calibrated null of \cref{prop:calibrated}, with a reference sample large enough that its band does not dominate the corpus's own sampling error, and reporting the exact-null and calibrated bounds side by side.

\paragraph{Using the method.}
We suggest reporting the upper end over a range of budgets, as in \cref{fig:grid}, rather than for one choice. A key holder can check the document-level condition on its own generations under edits it can simulate, using the normalized ratio of \cref{sec:experiments}; the relevant edits are those of real users, so this is a diagnostic, not a guarantee. A simple robustness device is a contamination share $\eta$: if a fraction $\eta$ of watermarked documents may fall outside the class but is still at least as detectable as human text, the upper end becomes $U/(1-\eta)$, and the smallest $\eta$ that moves the upper end past a decision threshold is a natural breakdown point \citep{horowitz1995identification}. A single threshold chosen without data to maximize $\hlow(t)-t$ is a reasonable default; in our simulations the multi-threshold bound was never far from the best single threshold, so it is a cheap guard against a poor choice. For monitoring, \cref{thm:running} needs no model of how adoption changes, but it still needs the exact null and the envelope conditionally on the past, and labels must not be chosen after seeing the document.

\paragraph{Adversaries.} The guarantees are not adversarial. A user who paraphrases deliberately can push documents outside the class and make the upper end fail; text crafted to look watermarked (spoofing) would bias the lower end. Calibration must be redone when the key, the tokenizer, the decoding rule or the model changes. Post-trained models produce lower-entropy text than the base models we use, which makes detection harder \citep{bahri2026entropy} and makes the entropy budget $\delta$ more consequential.

\paragraph{Limitations.}
The guarantees are relative to an edit model in which each position is retained, reset or substituted in place, independently given the past. LLM paraphrasing is outside that model, and the KGW paraphrase pools fall below the envelope at moderate thresholds; bounds that hold under paraphrasing would need a different restriction, for example on the semantic distance of the rewrite. Our random edits share token survivals between neighbouring positions, which the model does not describe exactly. The exact null is an assumption. Three keys and two batches of human text found no robust departure from it, but cannot rule out a key or domain effect below what twelve thousand documents detect, and the difference between our two human batches is unexplained. The calibrated null replaces this assumption with another: the reference sample must come from the same population as the human documents in the audited corpus, which is hard to guarantee when the corpus mixes domains. In our experiments corpora were drawn without replacement from finite pools, so coverage is conditional on those pools. The envelope must hold conditionally on length (\cref{ass:length}). Its numerical lower bound is conservative in exact arithmetic, and the floating-point error, checked against direct summation at selected lengths and lattice resolutions, was measured rather than proved small. Sharpness is with respect to the relaxed class. The experiments use two small open base models at temperature $1$, one context width, two schemes and synthetic mixtures. They do not cover instruction-tuned models, lower temperatures, low-entropy domains such as code, repeated-context masking at generation time, edits with a fixed count rather than a random share, or paraphrase pools larger than about 300 documents per model and scheme; these are the settings in which the envelope is most likely to fail. No corpus with known watermark prevalence is public, and deployed schemes such as SynthID \citep{dathathri2024synthid} keep their keys private. The method needs only per-document pivots, so it is available to any key holder.

\section*{Statements}
\paragraph{Code and data.} Code for the envelopes, intervals, confidence sequences, simulations and real-model experiments, together with all result files behind the tables and figures, is available at \url{https://github.com/YYKKCC-kevin/watermark-prevalence}. The experiments use public models and the public C4 validation split.
\paragraph{Ethics.} The work uses no human subjects or private data. The target is an aggregate, but the procedure computes a $p$-value for every document, and a key holder could use those to flag individuals. Aggregate estimates can also stigmatize a venue or community. We recommend reporting only the corpus-level interval and the budgets behind it.
\paragraph{Use of AI tools.} Large language model assistants (Claude and Codex) were used to help derive and check proofs, write and run simulation and experiment code, and draft text. The author checked all statements, proofs, numbers and references and takes full responsibility for the content.
\paragraph{Funding.} This work received no specific funding.
\paragraph{Author contributions.} K.Y.\ is the sole author and carried out the conceptualization, methodology, formal analysis, software, investigation, visualization and writing.
\paragraph{Competing interests.} The author declares no competing interests.

\bibliographystyle{plainnat}
\bibliography{refs}

\appendix
\section{Proofs}\label{app:proofs}

\paragraph{Proof of \cref{lem:lf}.} The result is due to \citet[Section~3.2]{li2025framework}; we give a short self-contained version.
Fix $x\in(0,1)$ and let $g(p)=p\,x^{1/p}$ for $p\in(0,1]$, with $g(0)=0$. Then $g'(p)=x^{1/p}(1-\ln x/p)>0$ and $g''(p)=x^{1/p}(\ln x)^2/p^3>0$, so $g$ is increasing and strictly convex, and it is continuous at $0$. The function $P\mapsto F_{1,P}(x)=\sum_wg(P(w))$ is convex on the polytope $\{P\ge0,\ \sum_wP(w)=1,\ P(w)\le c\}$ with $c=1-\delta$, so its maximum is attained at a vertex. At a vertex at most one coordinate lies strictly between $0$ and $c$. Hence the vertices are, up to permutation, $k=\lfloor1/c\rfloor$ coordinates equal to $c$, one coordinate equal to $r=1-kc\in[0,c)$, and zeros. For $\delta\in(0,\frac12)$ we have $k=1$ and $r=\delta$; for $\delta=\frac12$, $k=2$ and $r=0$. In both cases the vertex is $P^*=(1-\delta,\delta,0,\dots)$, and $F_{1,P^*}(x)=F^*(x;\delta)$. All vertices are permutations of each other, so they give the same value. When $|V|>2$, attainment uses zero-probability coordinates, which the closed simplex allows. If one insists on full support, $F^*$ is a supremum, and every statement below holds with ``attained'' replaced by ``approached''. \qed

\paragraph{Proof of \cref{lem:subst}.}
\emph{Gumbel-max.} Set $F_{\mathrm{sub}}(x;0)=x$ by continuity. Fix the context, the NTP $P$ and the emitted token $w$ with $p=P(w)$. Write $a=U_w^{1/p}$. The token $w$ wins exactly when $U_v<a^{P(v)}$ for every $v\neq w$. Since $a$ has density $p\,s^{p-1}$ on $(0,1)$ and, given $a$, the win has probability $\prod_{v\ne w}a^{P(v)}=a^{1-p}$, the joint density of $a$ and the win is $p$, so $a$ is uniform given the win. Given $a$ and the win, the $U_v$ are independent and uniform on $[0,a^{P(v)}]$. The replacement $w'$ depends on the text only, that is on $w$ and the history, and not on the key, so given the win its key value is $U_{w'}=a^qV$ with $q=P(w')$ and $V$ uniform and independent of $a$. Hence
\[
\Pr(U_{w'}\le x)=\int_0^1\min\{1,xa^{-q}\}\,da=x^{1/q}+\int_{x^{1/q}}^1xa^{-q}\,da=x^{1/q}+\frac{x(1-x^{1/q-1})}{1-q}.
\]
For fixed $x$, $a^q$ decreases in $q$, so $\min\{1,xa^{-q}\}$ and the integral increase in $q$; at $q\to0$ the value is $x$, so $F_{\mathrm{sub}}\ge x$. Under $\max_wP(w)\le1-\delta$, $q\le1-\delta$.

\emph{KGW.} Fix the context, $P$, the emitted token $w$ and whether $w$ is green. Before emission, and given this information about $w$, the green list is uniform over the $\gamma|V|$-subsets consistent with it; emission then reweights these lists, which the argument below accounts for. If every consistent list contains $w'$, or none does, the claim is immediate. Let $A$ be the consistent lists containing $w'$ and $B$ those that do not; a priori $w'$ is green with probability at least $g=(\gamma|V|-1)/(|V|-1)$. The emission probability of $w$ under list $\Gamma$ is $P(w)e^{b\1\{w\in\Gamma\}}/(1+cS(\Gamma))$. Pair each list in $A$ with the lists in $B$ obtained by swapping $w'$ for a token $u\notin\Gamma\cup\{w\}$. This bipartite graph is biregular, and along each edge the green mass of the list in $A$ exceeds that of its partner in $B$ by $q-P(u)\le q$. So the emission probability under the list in $A$ is at least $1/(1+cq)$ times that under its partner. Summing over edges and using biregularity, the posterior odds of $A$ against $B$ are at least the prior odds $|A|/|B|$ divided by $1+cq$, which gives $\Pr(w'\text{ green}\mid w\text{ emitted})\ge g/\{g+(1-g)(1+cq)\}=g_{\mathrm{sub}}(q)$. This is decreasing in $q$, and $q\le1-\delta$. Exhaustive enumeration over all green lists for $|V|=10$ and 300 random NTPs confirmed the bound, with a smallest margin of $0.020$ (\texttt{sim/substitution.py}), and Monte Carlo confirmed the Gumbel formula to within sampling error. \qed

\paragraph{Proof of \cref{lem:coupling}.}
If the past contains more than the previous pivots, for example the text and the NTPs, first condition down to the pivot history. By the tower property, the conditional distribution function of $Y_t$ given $Y_1,\dots,Y_{t-1}$ is an average of functions bounded by $\bar F$, so it is bounded by $\bar F$ too. Let $F_t(\cdot\mid Y_1,\dots,Y_{t-1})$ be that conditional distribution function and $F_t^{-1}$ its generalized inverse. Let $V_1,\dots,V_n$ be independent $U(0,1)$ variables. Defining $Y_t=F_t^{-1}(V_t\mid Y_1,\dots,Y_{t-1})$ recursively reproduces the joint law of $(Y_1,\dots,Y_n)$. Set $\bar Y_t=\bar F^{-1}(V_t)$, which are independent with law $\bar F$. Since $F_t\le\bar F$ pointwise, $F_t^{-1}\ge\bar F^{-1}$, so $Y_t\ge\bar Y_t$ for every $t$, and $T_n\ge\bar T_n$ almost surely because $\phi$ is nondecreasing. \qed

\paragraph{Proof of \cref{prop:envelope}.}
For a document in $\cA(\epsmin,\delta)$, the conditional distribution function of $Y_t$ given the past is $(1-\varepsilon_i-\sigma_i)x+\varepsilon_iF_{1,P_t}(x)+\sigma_iF_{s,t}(x)$. This uses that $P_t$ is a function of the past and $\zeta_t$ is fresh. By \cref{lem:lf}, $F_{1,P_t}\le F^*$, and by \cref{lem:subst}, $F_{s,t}\le F_{\mathrm{sub}}(\cdot;1-\delta)$. Since $F_{1,P}(x)\le x\le F_{\mathrm{sub}}(x;1-\delta)$, the expression decreases in $\varepsilon_i$ and increases in $\sigma_i$, so with $\varepsilon_i\ge\epsmin$ and $\sigma_i\le\sigma_{\max}$ it is at most $\bar F$ in \eqref{eq:Fbarsub}. \Cref{lem:coupling} with $\phi(y)=-\log(1-y)$ gives $T_n\ge_{\mathrm{st}}\bar T_n$, and $p=\bar\Gamma_n(T_n)$ is decreasing in $T_n$, so $\Pr(p\le t)\ge\Pr(\bar T_n\ge\bar\Gamma_n^{-1}(t))=h(t;n)$. With $\sigma_{\max}=0$, under $\varepsilon_i\equiv\epsmin$ and $P_t\equiv P^*$ the pivots are independent with law exactly $\bar F$, so equality holds.

For strictness, use the coupling with $\bar Y_t=\bar F^{-1}(V_t)$ and the null pivots $V_t$. Write $\phi(v)=-\log(1-v)$, $T^0_n=\sum_t\phi(V_t)\sim\Gamma(n,1)$, and $\bar T_n=T^0_n+\sum_tD(V_t)$ with $D(v)=\phi(\bar F^{-1}(v))-\phi(v)$. Because $\bar F$ is continuous, strictly increasing and, by the assumption of the strictness claim, satisfies $\bar F(x)<x$ on $(0,1)$, $D$ is continuous and strictly positive on $(0,1)$. Let $c=\bar\Gamma_n^{-1}(t)\in(0,\infty)$. Since $\sum_t\phi(v_t)$ is continuous on $(0,1)^n$ with range $(0,\infty)$, pick $v^*$ with $\sum_t\phi(v^*_t)=c$, and let $\eta=\frac12\sum_tD(v^*_t)>0$. By continuity there is an open ball $\mathcal O\subset(0,1)^n$ around $v^*$ on which $\sum_tD(v_t)>\eta$ and $\sum_t\phi(v_t)>c-\eta$. The set $\mathcal O^-=\{v\in\mathcal O:\sum_t\phi(v_t)<c\}$ is open and nonempty (lower one coordinate of $v^*$ slightly), so it has positive Lebesgue measure. On $\mathcal O^-$, $T^0_n<c$ and $\bar T_n=T^0_n+\sum_tD(V_t)>c-\eta+\eta=c$. Hence $h(t;n)-t=\Pr(\bar T_n\ge c)-\Pr(T^0_n\ge c)>0$. \qed

\paragraph{Proof of \cref{thm:idset}.}
\emph{Validity.} If $\pi_n=0$ the lower bound is trivial. If $\pi_n>0$, then $G_n=(F_n-(1-\pi_n)U)/\pi_n$ is absolutely continuous because $F_n$ is, with a density $g_n\ge0$, so $f_n=(1-\pi_n)+\pi_ng_n\ge1-\pi_n$ almost everywhere and $\pi_n\ge1-\essinf f_n$. For every $t$, $F_n(t)-t=\pi_n(G_n(t)-t)\ge\pi_n(h_n(t)-t)$, and $h_n(t)>t$, so $\pi_n\le(F_n(t)-t)/(h_n(t)-t)$; together with $\pi_n\le1$ this gives the upper end.

\emph{Sharpness.} If $F_n=U$, then $\pi_{n,L}=\pi_{n,U}=0$ and $\pi_n=0$ is the only decomposition, because $G_n\ge h_n>U$ forces $F_n>U$ somewhere whenever $\pi_n>0$. Otherwise $\essinf f_n<1$, so $\pi_{n,L}>0$. Take $\pi\in[\pi_{n,L},\pi_{n,U}]$ and set $G^\pi(t)=t+(F_n(t)-t)/\pi$. It is absolutely continuous with density $1+(f_n-1)/\pi$, which is nonnegative almost everywhere exactly when $f_n\ge1-\pi$, that is, when $\pi\ge\pi_{n,L}$. It satisfies $G^\pi(0)=0$ and $G^\pi(1)=1$, so it is a distribution function on $[0,1]$. Finally $G^\pi(t)\ge h_n(t)$ if and only if $(F_n(t)-t)/\pi\ge h_n(t)-t$, which holds for all $t$ when $\pi\le\pi_{n,U}$. Then $F_n=(1-\pi)U+\pi G^\pi$ with $G^\pi\in\Grel_n$. \qed

\paragraph{Proof of \cref{lem:kgw}.}
Write $u=1+cS$. Then $\psi(S)=\frac{c+2}{c}-\frac{c+1}{cu}-\frac uc$, and differentiating twice in $S$ gives $\psi''(S)=-2c(c+1)/(1+cS)^3<0$ on $[0,1]$. For a fixed green set, $S(P)$ is linear in $P$, so $P\mapsto\psi(S(P))$ is concave, and so is its expectation over $\Gamma$. A concave function on a polytope attains its minimum at a vertex. The polytope and its vertices are those in the proof of \cref{lem:lf}. Because $\Gamma$ is uniform, all permutations of $P^*$ give the same value. Under $P^*$, $S$ equals $1$ if both support tokens are green, $0$ if neither is, $1-\delta$ if only the first is, and $\delta$ if only the second is. The last two events each have probability $p_{10}$ under sampling without replacement, and $\psi(0)=\psi(1)=0$. \qed

\paragraph{Proof of \cref{prop:kgw}.}
Given the past, $\Gamma_t$ is uniform and $P_t$ is fixed, so under \eqref{eq:modelstar} the conditional probability that position $t$ is green is $(1-\varepsilon_i-\sigma_i)\gamma+\varepsilon_i\mu(P_t)+\sigma_i\,g_t$, where $g_t\ge g_{\mathrm{sub}}(1-\delta)$ is the green probability of the substituted token (\cref{lem:subst}). This equals $\gamma+\varepsilon_i(\mu(P_t)-\gamma)-\sigma_i(\gamma-g_t)\ge\gamma+\epsmin\Delta_{\mathrm{KGW}}(\delta)-\sigma_{\max}\{\gamma-g_{\mathrm{sub}}(1-\delta)\}=\bar\mu$ by \cref{lem:kgw}. \Cref{lem:coupling} with $\phi=\mathrm{id}$ gives $T_n\ge_{\mathrm{st}}\bar T_n\sim\mathrm{Bin}(n,\bar\mu)$. For fixed $U$, the randomized $p$-value is nonincreasing in $T$: $\bar B_n(k+1)+Ub_n(k+1)\le\bar B_n(k+1)+b_n(k+1)=\bar B_n(k)\le\bar B_n(k)+Ub_n(k)$. Hence $\Pr(p\le t)\ge\sum_k\Pr(\bar T_n=k)\,\Pr(\bar B_n(k)+Ub_n(k)\le t)$, which is the stated sum. Here $U$ is independent of the count and of the history. The function $k\mapsto\Pr(\bar B_n(k)+Ub_n(k)\le t)$ is nondecreasing, and for $t\in(0,1)$ it is not constant. $\mathrm{Bin}(n,\bar\mu)$ with $\bar\mu>\gamma$ strictly dominates $\mathrm{Bin}(n,\gamma)$ in the likelihood-ratio order, so $h_{\mathrm{KGW}}(t;n)>t$. \qed

\paragraph{Proof of \cref{prop:degenerate}.}
If $P_t$ is a point mass, \eqref{eq:F1} gives $F_{1,P_t}(x)=x$. For KGW, $S\in\{0,1\}$ and $\psi(S)=0$, so the green probability is $\gamma$. In both cases \eqref{eq:modelstar} reduces to $F_0$ for every $\varepsilon_i$, the pivots of a watermarked document have the null law, and its $p$-value is uniform. For any $\pi'\in[\pi,1]$, relabelling a fraction of human documents as such watermarked documents leaves $F_n$ unchanged and stays in the mechanism class. \qed

\paragraph{Proof of \cref{prop:reference}.}
(i) A single pivot has distribution function $(1-\pi)F_0+\pi\E_\Lambda[(1-\varepsilon)F_0+\varepsilon F_1]=F_0+\pi\E_\Lambda[\varepsilon](F_1-F_0)$. It depends on $(\pi,\Lambda)$ only through $\pi\E_\Lambda[\varepsilon]$ and determines that product because $F_1\neq F_0$.

(ii) If $\pi'=\pi$ there is nothing to show. Otherwise let $Q_\varepsilon=[(1-\varepsilon)F_0+\varepsilon F_1]^{\otimes n}$, so $Q_0=F_0^{\otimes n}$. Under $(\pi',\Lambda')$ with $\Lambda'=\frac{\pi}{\pi'}\Lambda+(1-\frac{\pi}{\pi'})\delta_0$, the pivot law is
\[
(1-\pi')F_0^{\otimes n}+\pi'\!\int Q_\varepsilon\,d\Lambda'(\varepsilon)
=(1-\pi')F_0^{\otimes n}+\pi\!\int Q_\varepsilon\,d\Lambda(\varepsilon)+(\pi'-\pi)F_0^{\otimes n},
\]
which is the law under $(\pi,\Lambda)$.

(iii) Because $F_1\neq F_0$ there is a bounded $u$ with $\E_{F_0}u=0$ and $d:=\E_{F_1}u\neq0$. Given $\varepsilon$, a watermarked document's pivots are independent with $\E[u(Y_j)\mid\varepsilon]=\varepsilon d$, and a human document's product has mean $0$. Hence, for $1\le k\le n$,
\[
\E\Big[\prod_{j\le k}u(Y_j)\Big]=d^k\,\pi\,\E_\Lambda[\varepsilon^k],
\]
so the pivot law determines $\{\pi\E_\Lambda[\varepsilon^k]\}_{k\le n}$. Conversely, expanding the product, the pivot law equals $(1-\pi)F_0^{\otimes n}+\pi\sum_{A\subseteq\{1,\dots,n\}}\E_\Lambda[\varepsilon^{|A|}(1-\varepsilon)^{n-|A|}]\bigotimes_{j}(F_1\text{ if }j\in A\text{, else }F_0)$. For $A\neq\emptyset$ the coefficient $\pi\E_\Lambda[\varepsilon^{|A|}(1-\varepsilon)^{n-|A|}]$ is a linear combination of $\pi\E_\Lambda[\varepsilon^j]$, $j\ge|A|$. The coefficient of $F_0^{\otimes n}$ is then fixed because the coefficients sum to $1$. So these $n$ quantities determine the law, and they are exactly what is identified. The bound follows from $\E_\Lambda[\varepsilon]\ge\epsmin$. \qed

\section{Relaxed class versus mechanism class}\label{app:relaxed}

The sharpness in \cref{thm:idset} is with respect to $\Grel_n$. The class of alternatives actually produced by $\cA(\epsmin,\delta)$ is the convex hull of mechanism-generated distributions, which is smaller. Without substitutions, if the data come from the least-favourable mechanism, both classes give the same upper end. In a linear-programming diagnostic over the convex hull of a dictionary of 42 distinct mechanisms that includes $P^*$ (\cref{app:lp}), which is an inner approximation of the mechanism class, the upper bound was only $0.004$ to $0.007$ below the relaxed one on the two data-generating processes we tried. We therefore state results for the relaxed class. The relaxation also discards restrictions from above. For KGW, $\varphi$ is concave, so by Jensen's inequality $\mu(P)\le\varphi(\gamma)<1$, and, without substitutions, watermarked documents of a fixed length cannot be detected with arbitrarily high power at a fixed threshold. An upper envelope built from this bound would raise the lower end of the identified set; we leave it to future work.

\section{Dictionary diagnostic for the mechanism class}\label{app:lp}

To gauge how much the relaxed class loses, we computed an upper bound over the convex hull of a finite dictionary of mechanism-generated alternatives. The dictionary has 42 distinct members: retention $\varepsilon\in\{0.3,0.4,0.5,0.6,0.8,1\}$ crossed with seven NTP shapes, namely top probability $q\in\{0.3,0.5,0.7,0.8\}$ with the remainder spread over nine tokens, and $q\in\{0.5,0.7,0.8\}$ with the remainder on one token. The last of these is $P^*=(0.8,0.2)$. Each member's $p$-value distribution is computed by FFT convolution on a fine grid, without Monte Carlo error but with discretization error that we did not bound. With variables $v_j=\pi w_j$ the program is
\[
\max\ \pi\quad\text{s.t.}\quad\Big|(1-\pi)t_k+\sum_jv_jG_j(t_k)-\hat F(t_k)\Big|\le b\ \ \forall k,\qquad\sum_jv_j=\pi\le1,\qquad v\ge0 .
\]
With $n=500$, $N=20{,}000$, $\pi=0.2$, $b=0.0096$ and 100 replications, the LP gave $[0.192,0.213]$ (coverage $0.99$) against $[0.192,0.220]$ for the relaxed class (coverage $1.00$) on an in-class process with varying $q_t$. On the least-favourable process it gave $[0.186,0.209]$ against $[0.183,0.213]$. A dictionary is an inner approximation of the mechanism class, so the LP has no coverage guarantee over the whole class. Two processes also cannot show that the relaxed class is nearly lossless in general. An earlier dictionary that omitted $P^*$ appeared to be much tighter; that gap was an artefact of the missing member.

\section{Additional results and implementation details}\label{app:extra}

\begin{figure}[h]
\centering
\includegraphics[width=\linewidth]{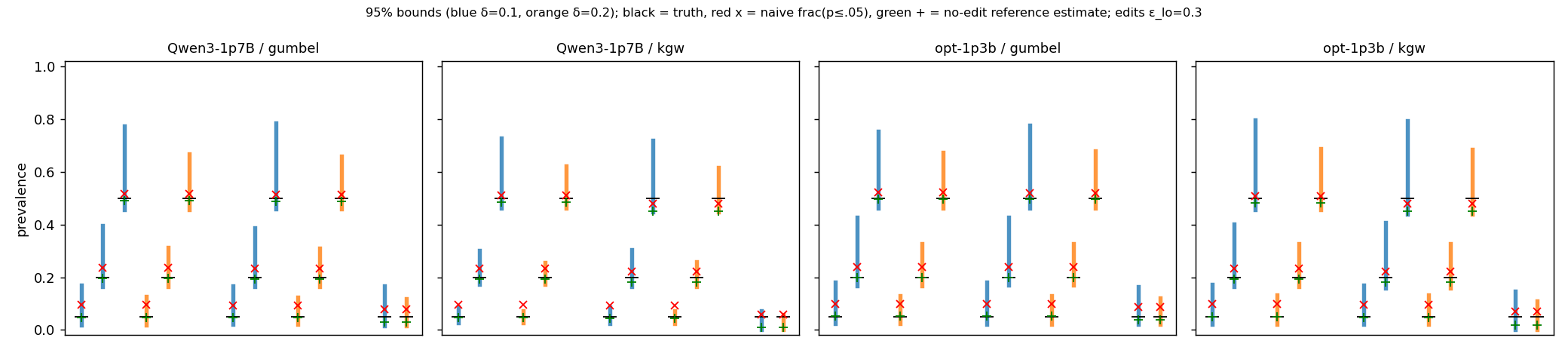}
\caption{Fixed-composition design. Average 95\% intervals for every pool and prevalence ($\varepsilon_{\mathrm{lo}}=0.3$; blue $\delta=0.1$, orange $\delta=0.2$). Black ticks mark the truth, red crosses the detection rate, and green plus signs the no-edit reference estimate.}
\label{fig:bounds}
\end{figure}

\paragraph{Fixed-composition design with real human $p$-values.}
Before the design of \cref{sec:experiments}, we evaluated a second design. Each corpus mixes exactly $\pi N$ watermarked documents, drawn without replacement from a pool, with real human documents, and the mixing is repeated 50 times. The replications are heavily overlapping subsamples of fixed pools; at $\pi=0.05$ each corpus uses 3800 of the 4000 human documents. Coverage is therefore close to conditional on the pools, and its target, the realized fraction, differs from $\pi_{\mathrm{cond}}$. In this design coverage was $1.00$ for every random-edit cell. For paraphrases, which were evaluated only at $\pi=0.05$, it was $0.90$ for Qwen3 with KGW at $\delta=0.2$ and $1.00$ otherwise. Across span edits and $\varepsilon_{\mathrm{lo}}=0.5$, the average interval endpoints differ from the scattered, $\varepsilon_{\mathrm{lo}}=0.3$ entries by at most $0.015$ for the lower end and at most $0.023$ ($\delta=0.1$) and $0.007$ ($\delta=0.2$) for the upper end, over all models, schemes and prevalences. The two Gumbel-max scattered pools at $\varepsilon_{\mathrm{lo}}=0.5$ have 1991 and 1997 documents, fewer than the 2000 needed for $\pi=0.5$ at $N=4000$, so that cell is missing. \Cref{tab:extra} gives the scattered, $\varepsilon_{\mathrm{lo}}=0.3$ results of this design, and \cref{fig:bounds} shows all configurations.

\begin{table}[h]
\centering
\caption{Fixed-composition design with real human $p$-values. Random scattered edits, $\varepsilon_{\mathrm{lo}}=0.3$, $N=4000$, 50 overlapping replications; average 95\% interval endpoints and point estimates. Coverage was $1.00$ in all cells.}
\label{tab:extra}
\begin{tabular}{llcccc}
\toprule
$\pi$ & model / scheme & $\delta=0.1$ & upper, $\delta=0.2$ & detection rate & null-corrected \\
\midrule
0.05 & OPT / Gumbel   & $[0.023,0.182]$ & 0.131 & 0.099 & 0.051 \\
     & Qwen3 / Gumbel & $[0.018,0.170]$ & 0.127 & 0.094 & 0.047 \\
     & OPT / KGW      & $[0.022,0.174]$ & 0.133 & 0.098 & 0.051 \\
     & Qwen3 / KGW    & $[0.027,0.086]$ & 0.072 & 0.096 & 0.048 \\
\midrule
0.5  & OPT / Gumbel   & $[0.461,0.753]$ & 0.675 & 0.522 & 0.497 \\
     & Qwen3 / Gumbel & $[0.457,0.775]$ & 0.667 & 0.515 & 0.490 \\
     & OPT / KGW      & $[0.455,0.797]$ & 0.689 & 0.508 & 0.482 \\
     & Qwen3 / KGW    & $[0.462,0.728]$ & 0.623 & 0.510 & 0.484 \\
\bottomrule
\end{tabular}
\end{table}

\paragraph{Implementation.}
Real-model design (\cref{sec:experiments}): strata of scored length $[50,100)$, $[100,200)$, $[200,\infty)$; thresholds $\{0.01,0.02,0.05,0.1,0.2,\dots,0.9\}$; $\alpha=0.05$ for \cref{prop:realized}, $\alpha=\alpha_{\mathrm{ref}}=0.025$ for \cref{prop:calibrated} with a DKW band per stratum at level $\alpha_{\mathrm{ref}}/K$; reference split of 4000 and audit split of 8000 fresh human documents, redrawn every 50 corpora. Weak-signal simulations: the same twelve thresholds are the candidates for every single-threshold rule; the pilot corpus has the same size and prevalence as the audited one, is drawn independently in every replication, and the pilot rule picks the threshold with the smallest single-threshold upper bound on it. Threshold grid for \cref{cor:band}: $t\in\{0.001,0.005\}\cup\{0.01,0.02,\dots,0.99\}$. Conservative envelope: lattice step $0.01$, truncation $S_{\max}=30$, tolerance $10^{-8}$; the floating-point comparison is in \texttt{sim/check\_fft\_error.py}. Betting bets: $\lambda_i=\min\{\sqrt{2\log(2/\alpha)/(\hat\sigma^2_{i-1}\,i\log(1+i))},\frac12\}$ with $\hat\sigma^2_{i-1}$ the running variance of $B_1,\dots,B_{i-1}$ regularized toward $\frac14$, computed on a $\pi_0$ grid of step $0.002$ with outer approximation. Realized-fraction sequence: $\rho=N/40$, with $N$ the planned horizon, fixed before the stream starts. Watermark context: SHA-256 of the key and the previous three token identifiers. All GPU work ran on one RTX A6000, with several jobs sharing it at once: about 1.7 hours of wall-clock time and about 5 hours of summed job time for the original experiments, and about 1.3 hours of wall-clock time for the calibration study.

\section{A betting confidence sequence for a fixed prevalence}\label{app:betting}
Documents arrive one at a time; within a stratum fix $t$ and let $B_i=\1\{p_i\le t\}$. 
Let $\cF_i=\sigma(p_1,\dots,p_i)$. Assume that $\Pr(Z_i=1\mid\cF_{i-1})=\pi$, that $p_i$ is uniform given $\cF_{i-1}$ and $Z_i=0$, and that given $\cF_{i-1}$ and $Z_i=1$ the document lies in $\cA(\epsmin,\delta)$. The mechanism may depend on the past, so the edits can be chosen adaptively. Then the conditional mean $m_i=\E[B_i\mid\cF_{i-1}]$ satisfies the following bounds whenever $\hlow(t)\le h(t)$; because that inequality is established only up to floating-point error, we state the bounds themselves as the assumption of \cref{thm:betting}:
\begin{equation}\label{eq:mbounds}
\mu_L(\pi):=t+\pi\,(\hlow(t)-t)\ \le\ m_i\ \le\ t+\pi(1-t)=:\mu_U(\pi).
\end{equation}
For thresholds $t_1<\dots<t_J$ and predictable bets $\lambda_{ij}\in[0,\frac12]$, define the two capital processes
\[
K^{\uparrow}_{k,j}(\pi_0)=\prod_{i\le k}\big(1+\lambda_{ij}(\mu_{L,j}(\pi_0)-B_{ij})\big),\qquad
K^{\downarrow}_{k,j}(\pi_0)=\prod_{i\le k}\big(1+\lambda_{ij}(B_{ij}-\mu_{U,j}(\pi_0))\big),
\]
and the averaged capital $K_k(\pi_0)=\frac1{2J}\sum_j\big[K^{\uparrow}_{k,j}(\pi_0)+K^{\downarrow}_{k,j}(\pi_0)\big]$.

\begin{theorem}[Betting confidence sequence]\label{thm:betting}
Under the assumptions above, $C_k=\{\pi_0\in[0,1]:K_k(\pi_0)<1/\alpha\}$ satisfies $\Pr(\pi\in C_k\text{ for all }k\ge1)\ge1-\alpha$, and so does the running intersection $\bigcap_{k'\le k}C_{k'}$. If the bets do not depend on $\pi_0$, $K_k$ is convex in $\pi_0$ and $C_k$ is an interval.
\end{theorem}

\begin{proof}
Because $\hlow$ is clipped to $[0,1]$, $\mu_L$ and $\mu_U$ lie in $[0,1]$, and with $\lambda_{ij}\le\frac12$ every factor is at least $\frac12$. At $\pi_0=\pi$, \eqref{eq:mbounds} gives each factor of $K^{\uparrow}$ and $K^{\downarrow}$ conditional mean at most $1$, so each is a nonnegative supermartingale with initial value $1$. Their average is too, and Ville's inequality \citep{ville1939etude} gives $\Pr(\exists k:K_k(\pi)\ge1/\alpha)\le\alpha$. Each factor of $K^\uparrow_{k,j}$ is positive and affine in $\pi_0$ with slope $\lambda_{ij}(\hlow(t_j)-t_j)$, whose sign is the same for all $i$; each factor of $K^\downarrow_{k,j}$ is positive and affine with slope $-\lambda_{ij}(1-t_j)\le0$. If $f=\prod_i(a_i+b_i\pi_0)$ with positive factors and slopes of one sign, then $f''=f\sum_{i\ne i'}c_ic_{i'}\ge0$ with $c_i=b_i/(a_i+b_i\pi_0)$, so each capital process is convex, and so is their average.
\end{proof}

The construction follows the hedged capital process of \citet{waudbysmith2024estimating}. We use their predictable plug-in bets. Averaging over thresholds needs no multiplicity correction for validity, but it is not free: when one component alone drives rejection, it must reach about $2J/\alpha$. In return the data choose the most informative threshold. When computing $C_k$ on a grid of $\pi_0$ values we take an outer approximation. Because $C_k$ is an interval, extending each end of the accepted grid points to the adjacent rejected grid point contains it. If no grid point is accepted at time $k$, the interval $C_k$ may still be nonempty between two grid points; we then skip the intersection at time $k$, which only enlarges the running interval and keeps it valid. The running interval starts at $[0,1]$.

\paragraph{Simulations of both confidence sequences.}
\Cref{tab:cs} summarizes both sequences over $N=20{,}000$ arriving documents and 50 replications, recording whether the interval covered the target at \emph{every} time $k$. The betting sequence of \cref{thm:betting} averages over $t\in\{0.01,0.05,0.1,0.2,0.5\}$. To test the outer grid approximation, we set the true value to $0.2013$, between grid points. With $\pi=0$ it never reported a positive lower bound. At $k=20{,}000$ and $n=500$ its average interval was $[0.187,0.242]$. The fixed-sample interval of \cref{cor:band} in a comparable setting (a separate simulation with $\pi=0.2$) was $[0.191,0.218]$. The lower ends agree; the betting interval is about twice as wide, which is the cost of being valid at every $k$. The realized-fraction sequence of \cref{thm:running} covered $\bar\pi_k$ at all times under a drift ($0\to0.4\to0.1$), periodic bursts, and independent labels. It is about twice as wide as the betting sequence in the stationary case.

\begin{table}[t]
\centering
\caption{Confidence sequences, $N=20{,}000$ documents, 50 replications, assumed $(\epsmin,\delta)=(0.3,0.2)$. ``Coverage'' is the fraction of replications in which the target was inside the interval at every $k$. Betting intervals are averaged over replications; realized-fraction intervals are from one representative run.}
\label{tab:cs}
\begin{tabular}{llccc}
\toprule
sequence & target & $n$ & coverage & interval at $k=20{,}000$ \\
\midrule
betting (\cref{thm:betting}) & $\pi=0$       & 500 & 1.00 & $[0.000,0.008]$ \\
                             & $\pi=0.05$    & 500 & 1.00 & $[0.042,0.075]$ \\
                             & $\pi=0.2013$  & 500 & 0.98 & $[0.187,0.242]$ \\
                             & $\pi=0.2013$  & 200 & 1.00 & $[0.187,0.277]$ \\
\midrule
realized fraction (\cref{thm:running}) & drift $0\to0.4\to0.1$ & 500 & 1.00 & $[0.212,0.316]$, $\bar\pi_k=0.224$ \\
                                       & periodic bursts       & 500 & 1.00 & $[0.311,0.449]$, $\bar\pi_k=0.320$ \\
                                       & i.i.d.\ $0.2$         & 500 & 1.00 & $[0.197,0.296]$, $\bar\pi_k=0.207$ \\
\bottomrule
\end{tabular}
\end{table}

\end{document}